\documentclass{article}
\usepackage{ijcai20}

\usepackage{times}

\usepackage{soul}
\usepackage{url}
\usepackage[hidelinks]{hyperref}
\usepackage[utf8]{inputenc}
\usepackage[small]{caption}
\usepackage{graphicx}
\usepackage{amsmath}
\usepackage{booktabs}
  \usepackage[dvipsnames]{xcolor}	% allow ForestGreen color name
\usepackage{colortbl}
\usepackage{multirow}
\usepackage{fancyvrb}

\newcommand{\eat}[1]{}
\newcommand{\red}[1]{\textcolor{red}{#1}}

\newcommand{\green}[1]{\textcolor{ForestGreen}{#1}}

\usepackage{quoting}
\newenvironment{myquote}{                   % list without par spacings
  \parskip 0mm \begin{quoting}[vskip=0mm,leftmargin=2mm]}{
\end{quoting}}
\newenvironment{ite}{                     % list without par spacings
     \parskip 0cm \begin{itemize} \parskip 0cm \parsep 0cm \itemsep 0cm \topsep 0cm}{
        \end{itemize}} %  \parskip 0cm}
\newenvironment{enu}{                   % list without par spacings
     \parskip 0cm \begin{list}{}{\parsep 0cm \itemsep 0cm \topsep 0cm}}{
       \end{list}} %  \parskip 0cm}
\newenvironment{des}{                 % list without par spacings
     \parskip 0cm \begin{list}{}{\parsep 0cm \itemsep 0cm \topsep 0cm}}{
       \end{list}} %  \parskip 0cm}
\newcommand{\vt}{$\vert$}

\newcommand{\demourl}{https://rule-reasoning.apps.allenai.org/}
\newcommand{\dataurl}{https://allenai.org/data/ruletaker}

\newcommand{\DMax}{DMax}
\newcommand{\MMax}{MMax}

\title{Transformers as Soft Reasoners over Language}
\author{
Peter Clark, Oyvind Tafjord, Kyle Richardson \\
\affiliations
Allen Institute for AI, Seattle, WA
\emails
\{peterc,oyvindt,kyler\}@allenai.org
}

\begin{document}

\maketitle

% \footnote{[To put somewhere] By rule, we mean an explicit, general statement of facts that can be concluded from other facts.}

% \title{Can Language Models Reason with Rules?}

\begin{abstract}
Beginning with McCarthy’s Advice Taker (1959), AI has pursued the goal of
providing a system with explicit, general knowledge and having the system
reason over that knowledge. However, expressing the knowledge in a formal (logical
or probabilistic) representation has been a major obstacle to this research.
This paper investigates a modern approach to this problem where the facts and rules
% are provided as natural language sentences, thus bypassing a formal representation,
are provided as natural language sentences, thus bypassing a formal representation.
We train transformers to reason (or emulate reasoning) over these sentences using
synthetically generated data.
% Hmmm.....RuleTaker in the abstract? 
% Our RuleTaker models
% We
Our models, that we call RuleTakers,
provide the first empirical demonstration that this kind of
soft reasoning over language is learnable, can achieve high (99\%) accuracy,
and generalizes to test data requiring substantially deeper chaining than seen during training (95\%+ scores).
We also demonstrate that the models transfer well to two hand-authored rulebases, and
to rulebases paraphrased into more natural language. These findings are significant as it
suggests a new role for transformers, namely as limited ``soft theorem provers'' operating
over explicit theories in language. This in turn suggests new possibilities for
explainability, correctability, and counterfactual reasoning in question-answering.\footnote{
% Datasets are at \dataurl~and a live demo is available at~\demourl}
A live demo and all datasets are available at \demourl~and \dataurl}
\end{abstract}

\section{Introduction}

% Example 115-3 in att-noneg-2
\begin{figure}[t]
\centerline{
 \fbox{%
   \parbox{0.90\columnwidth}{
     {\small
{\it (Input Facts:)} Alan is blue. Alan is rough. Alan is young. \\
Bob is big. Bob is round. \\
Charlie is big. Charlie is blue. Charlie is green. \\
Dave is green. Dave is rough.
\vspace{1mm} \\
{\it (Input Rules:)} Big people are rough. \\ % - perhaps avoid this alternative syntax?
If someone is young and round then they are kind. \\
% If someone is big then they are rough. \\
If someone is round and big then they are blue. \\
% If someone is rough then they are green. \\
All rough people are green. \\
\vspace{1mm} \\
Q1: Bob is green. True/false? {\bf [Answer: T]} \\
Q2: Bob is kind. True/false? {\bf [F]} \\
Q3: Dave is blue. True/false? {\bf [F]}
}}}
}   % end small
% \vspace{-1mm}
\caption{Questions in our datasets involve reasoning with rules. The inputs to
the model are the context (facts + rules) and a question. The output is the T/F answer to the question.
Here the underlying reasoning for the true fact (Q1) is: Bob is big, therefore
rough (rule1) therefore green (rule4).
% (This is an example of depth 2 reasoning).
Note that the facts + rules themselves change for different questions in the datasets.}
% Facts not provably true are assumed false.}
% IJCAI final (closed-world assumption).}
\label{example}
\vspace{-2mm}
\end{figure}

% ======================================================================
%        NEW INTRODUCTION
% ======================================================================

AI has long pursued the goal of giving a system explicit {\it knowledge}, and
having it {\it reason} over that knowledge to reach conclusions, dating back to the
earliest years of the field, e.g., McCarthy's Advice Taker (\citeyear{Mccarthy1959ProgramsWC}),
and Newell and Simon's Logic Theorist (\citeyear{Newell1956TheLT}).
While this has resulted in impressive applications (e.g., \cite{metaxiotis2002expert}),
building and reasoning over the required formal representations has also proved
challenging \cite{musen1988brittleness}. In this work, we explore a modern approach to this goal,
and ask whether transformers can be trained to reason (or emulate reasoning)
using rules expressed in language, thus bypassing a formal representation.
If so, new opportunities for question-answering,
explainability, correctability, and counterfactual reasoning may become possible.

This goal is quite distinct from question-answering as selecting an answer span in a passage,
today's prevailing paradigm, e.g., \cite{Rajpurkar2016SQuAD10}. Rather,
%
% which wording is best?
% we want the system to reason over {\it all} the provided information to find
% we want the system to reason over all the provided rules to find
we want the system to reason over the provided rules to find
% we want the system to reason over the provided information to find
% 
conclusions that follow. Our goal is also distinct from that of {\it inducing}
rules from examples, e.g., given instances of family relationships,
inducing that a parent's parent is a grandparent \cite{clutrr},
% or inducing that `taller' is a transitive relation \cite{richardson2019probing},
something that transformers are already known to do well. Rather, here we provide
rules explicitly, and wish transformers to draw appropriate
conclusions, as illustrated in Figure~\ref{example}. Here, rather than inducing rules
from examples, our task involves learning to emulate a reasoning {\it algorithm}.

\begin{figure}[t]
\centering
    \includegraphics[width=0.7\columnwidth]{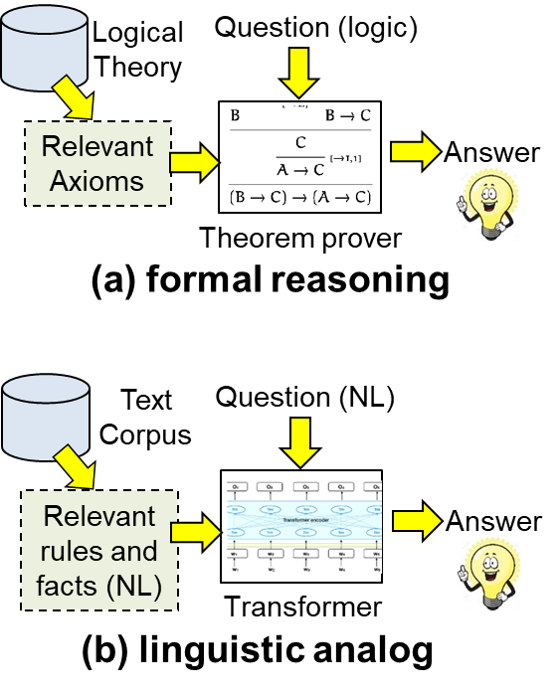}
    \vspace{-2mm}
    \caption{(a) Traditional formal reasoning applies a theorem prover to
    axioms in order to answer a question. (b) Our work here strives
    for a linguistic analog, where a transformer serves as a ``soft
    theorem prover'' over knowledge expressed linguistically.
%     (Our experiments  primarily use synthetic language).
    \label{logic-and-transformers}}
\end{figure}

% Our models, that we call RuleTakers, 
We provide the first demonstration that this is possible,
i.e., that transformers can reason with rules expressed in language.
Our approach uses a broadly applicable training regimen: Characterize the
desired behavior in a formal way, synthesize formal examples, generate linguistic
equivalents, and train a model. The result suggests a new role for transformers, namely
as a kind of limited ``soft theorem prover'' over language (Figure~\ref{logic-and-transformers}).
% IJCAI final
% \footnote{
% Strictly speaking, the model is not a theorem prover
% as it only outputs the proof results, not the proof itself. In Section~\ref{explanation},
% we describe first steps to also recovering the chain of reasoning.}
This in turn may allow inspection and control of the knowledge that the model
is manipulating, with potential benefits for explanation,
correctability, and counterfactual reasoning.

% ======================================================================

Our investigations here are in a limited setting:
Rules are linguistic expressions of conjunctive implications
{\it condition [$\wedge$ condition]* $\rightarrow$ conclusion},\footnote{
where {\it conditions} and {\it conclusions} are (possibly negated) literals using binary predicates.}
with the semantics of logic programs with negation  \cite{Apt1988TowardsAT};
and reasoning is the deduction of a statement's truth
according to these semantics.\footnote{We only consider rulebases where all statements have a well-defined truth value, i.e., consistent, stratified rulebases.}
However, although there is still a potentially large gap to natural
language inference (NLI),\footnote{
NLI is informally defined as making inferences from language that ``a person would typically infer'' \cite{Dagan2013RecognizingTE},
and includes use of many linguistic forms, unstated background knowledge, and sometimes unsound inference steps.}
our approach also suggests a path to teaching machines to reason over broader
language, with similar potential benefits. % for explanation, correction, and handling counterfactuals.

We leave open the question of whether the transformer is actually
``reasoning'', and even what that might mean in a neural setting.
Rather, we show that transformers can reliably emulate the i/o
behavior of a formal reasoner, including applied to test data
requiring more reasoning than at training time, two hand-authored rulebases,
and rulebases rephrased into more natural (crowdsourced) language. 

The paper is organized to address the following questions,
and contributes the following results:
\begin{enu}
\item[1.] {\bf Can transformers learn to reason with rules?}
We train and test on rules expressed in (synthetic) language,
and find high (99\%) accuracy, including on test questions
requiring a greater depth of reasoning than seen during training
(scoring up to 95\%, Table~\ref{main-results}).
% i.e., questions outside the scope of the training
% data, in particular generated at a greater depth of inference.} data,
% to test the generalizability and robustness of the model.
\item[2.] {\bf Can the trained model solve hand-authored reasoning problems?}
We find the trained models are able to solve five of six variants of
two independently authored rule-based problems, zero shot (90\%+ scores, Table~\ref{handcrafted}).
% We test on two rule-based puzzles (``birds'' and ``electricity'').
\item[3.] {\bf Do the results transfer to theories expressed in more natural language?}
Models also perform well when trained and tested on theories
paraphrased into more natural (crowdsourced) language (98\% score). The best earlier model
can even partially solve these problems zero-shot (66\% accuracy, Table~\ref{natlang}).
% The train models partially solve (50\% F1) theories expressed in natural language
% zero shot, and almost perfectly (95\%+ F1) when natural language data is
% added to training.
\item[4.] {\bf Can the model identify which facts an answer depends on?}
We show that the model is largely able to do this (94\% F1), including
perfect identification for over 70\% of the questions.
This is a first step towards having a model create an explanation
for its conclusions. (Section~\ref{explanation} and Figure~\ref{critical-histogram}).
% and may indicate that some kind of iterative computation is indeed going on internally. - bit assumptive
\item[5.] {\bf Can other neural architectures learn to reason?}
Our experiments show a particular transformer (RoBERTa) is sufficient for our tasks,
but is it necessary? We show that two other systems, BERT and ESIM (an LSTM-based model) \cite{chen2016enhanced},
are also able to learn these tasks, albeit with lower scores (95\% and 80\%
respectively, vs. 98\%). This suggests that our results are not
specific to RoBERTa or transformers, although transformers learn the tasks
more easily (Table~\ref{other-architectures}).
\end{enu}
Although our demonstrations are within constrained environments,
the work suggests new avenues for using transformers for
both formal theorem proving and natural language inference (NLI),
discussed in Section~\ref{discussion}.
In particular, the ability to derive reasoned conclusions from
stated knowledge suggests new opportunities for explainability,
correctability/machine instruction, and counterfactual reasoning
in question-answering.

\section{Related Work}

% The key element that distinguishes this work from the prior body of work
% is the systematic study of multiple rules and the use of transformers.
% However, there are datasets that make a first step in particular situations...

% OLD
% While there has not been a systematic study of transformers
% reasoning with explicitly stated rulebases, there are several
% datasets that make a first step towards this by testing whether neural
% NEW
% A key contribution of our work is the first systematic study
% of transformers reasoning with explicitly stated rulebases.
% However, there are datasets that make a first step towards
% this by testing whether neural
%
While our work is, to the best of our knowledge, the first systematic study of
transformers reasoning over explicitly stated rule sets, there are several
datasets that make a first step towards this by testing whether neural
systems can apply explicit, general knowledge in a particular situation.
Two synthetic datasets that test whether a single rule can be
applied correctly are as follows:
\begin{enu}
\item[1.] Task 15 in the bAbI dataset \cite{bAbI} applies 
rules of the form ``Xs are afraid of Ys'' to an instance, e.g.,
``Sheep are afraid of wolves. Gertrude is a sheep. What is Gertrude afraid of? A:wolves''
\item[2.] The synthetic, conditional probes in \cite{richardson2019probing} test single
rule application, e.g., ``If Joe has visited Potsdam then Anne has visited Tampa. Joe has visited Potsdam. Has Anne visited Tampa? A:yes''
\end{enu}
The associated papers also show that neural systems can learn these tasks almost perfectly (100\% and $>$97\% accuracy respectively).

Similarly, two reading comprehension datasets involve applying general rule(s) to text:
\begin{enu}
\item[3.] In the QuaRTz dataset \cite{Tafjord2019QuaRTzAO}, the ``rules'' are qualitative
relationships, e.g., ``A sunscreen with a higher SPF protects the skin longer.'',
and the dataset tests whether a system can apply a rule correctly
to a situation, e.g., ``Billy is wearing sunscreen with a lower SPF than Lucy. Who will be best protected from the sun?'',
thus testing one-step rule application.
\item[4.] ROPES (Reasoning over Paragraph Effects in Situations) \cite{Lin2019ReasoningOP} is
similar, except the general knowledge is contained within a paragraph-length
passage and must be applied to a paragraph-length situation.
\end{enu}
While the non-synthetic nature of QuaRTz and ROPES adds realism to the task, it
also complicates them as a test of ``machine reasoning'', in the sense of
(producing the i/o behavior of) chaining rules to reach a conclusion.
Rather, they test reading comprehension, requiring not just ``reasoning''
but also use of background knowledge and handling the diversity of language.

Although our core datasets may seem similar to the bAbI dataset \cite{bAbI}
in using synthetic data, our probes are qualitatively different.
Specifically, apart from bAbI Task 15 (above), the underlying rules needed to
infer an answer in the bAbI tasks are {\it implicit}.
% while our concern here is reasoning with explicit rule sets,	% IJCAI shortening
% potentially different for each example (Figure~\ref{example}).
For example, answering Task 1 questions such as ``Mary went to the hallway.
John went to the office. Where is Mary? A: hallway'' requires
inducing state-change rules such as ``X moves to Y $\rightarrow$ X at Y''.
In other words, the bAbI tasks test whether a system can learn and apply these underlying
rules from examples, while our concern here is reasoning with
{\it explicit} rule sets, potentially different for each example (Figure~\ref{example}).

Similarly, our work is qualitatively distinct from work on multihop
reasoning datasets, e.g., HotpotQA \cite{yang2018hotpotqa}. Again, for
those problems, the implicit rules of inference (i.e., which multihop chains
are valid) need to be inferred from examples (e.g., that ``causes''
is transitive). In our case, rules of inference are explicitly stated.

Our approach contrasts with prior efforts that attempt to
semantically parse language into a formal form, so that a formal
reasoner can then be applied \cite{semantic-parsing}. Despite
substantial research, semantic parsing remains challenging,
with few examples of systems that can reliably convert 
multi-sentence
text into formal theories. Instead, we explore
reasoning with language directly,
bypassing the semantic parsing task.

% Our work is closely related to Natural Logic \cite{MacCartney2014NaturalLA} and Natural Language Inference \cite{
% Our work can be seen as evaluating transformers for Natural Logic \cite{MacCartney2014NaturalLA} and Natural Language Inference \cite{Manning2009NaturalLI}
% Our work can be seen as evaluating transformers for (a subset of) Natural Logic \cite{MacCartney2014NaturalLA}.
Our work can be seen as evaluating transformers for (a subset of) Natural Logic \cite{MacCartney2014NaturalLA,moss2010natural},
i.e., formal inference over statements expressed in language.
It is also related to textual entailment and Natural Language Inference (NLI) \cite{Manning2009NaturalLI},
but with the important difference that NLI also allows
{\it unsupported} inferences that ``a person would typically infer'' \cite{Dagan2013RecognizingTE}.
We discuss bridging the gap between our work and NLI in Section~\ref{nli}.

% but with the important difference that NLI also allows for the use of
% {\it unstated} background knowledge to make inferences that ``a person
% would typically infer'' \cite{Dagan2013RecognizingTE}.
% We discuss this further in Section~\ref{nli}.
% {\it unstated} background knowledge to be used for inferences
% that ``a person would typically infer'' 

Several researchers have developed methods for Neural Theorem Proving (NTP),
combining symbolic and neural methods to reason step-wise over language-derived structures,
e.g., \cite{Weber2019NLPrologRW,Minervini2019DifferentiableRO,Minervini2018TowardsNT}.
Similarly, there has been work on SAT solving \cite{Selsam2018LearningAS},
approximate (DNF) model counting \cite{Abboud2019LearningTR}, and
formula embedding \cite{Abdelaziz2020AnES,Crouse2019ImprovingGN} with neural networks,
to help solve formal reasoning problems. While our goals are similar, we do not impose any
structure on the neural reasoning process, instead wanting to know if the (i/o of the) reasoning
process itself is learnable, using knowledge expressed in language.

% In this sense, our work is more analogous to recent work
% on approximate DNF (model) counting \cite{Abboud2019LearningTR}. In that work,
% a graph neural network (GNN) was used to replicate the i/o behavior
% of a DNF counting algorithm using a large collection of synthesized problem
% instances. Here we replicate 
% Rather, we wish to know how transformers perform with training data alone.

Our task can perhaps best be viewed as one of {\it algorithm emulation},
here for systematic reasoning with rules. There have been numerous
other demonstrations that transformers either already know \cite{Talmor2019oLMpicsO} or can learn % can learn (or emulate)
to emulate other algorithms, including for semantic parsing \cite{He2019EstablishingSB},
machine translation \cite{Wang2019LearningDT},
symbolic integration \cite{Lample2019DeepLF},
% approximate (DNF) model counting \cite{Abboud2019LearningTR},
and mathematics \cite{Saxton2019AnalysingMR}.
Here we investigate a transformer's ability to learn rule-based reasoning.

\section{Dataset Generation}

% \subsection{Overview}

To investigate a transformer's ability to emulate rule-based reasoning,
we generate five datasets requiring various depths of inference
to answer the questions, as we now describe.
Each example in a dataset is a
triple {\it(context,statement,answer)}, where
{\it context} has the form {\it (fact*,rule*)}, {\it statement}
is the question, namely a declarative sentence to prove, and
{\it answer} is either T (true) if {\it statement} deductively follows from
the context, or F if it does not (false under a closed-world
assumption, CWA). Facts, rules, and the question statements are expressed
in (synthetic) English.
% Figure~\ref{example} shows four questions against the same context.
Each example is essentially a (linguistic) standalone logical theory
with an ``Is it true?'' question (the {\it statement}) posed against it.
By virtue of the dataset generation procedure (below), we know every
question can be answered by a formal reasoner (under a CWA). We are
interested in whether a transformer can similarly learn to
answer these questions, apparently requiring some
neural form of reasoning.

% \subsection{Dataset Generation}

\subsection{Overview}

To generate each example, we first generate a small theory
(facts + rules) in logic, perform forward inference to derive 
all its implications, then select question statements from those
implications (answer=true), and from unproven (positive) facts
(answer=false, under the CWA). 

The facts, rules, and questions are then expressed in (synthetic) English using
simple natural language templates. We generate five datasets, each constrained by the
maximum depth of inference required to prove the facts used in its questions (up to depths
D=0, D$\leq$1, D$\leq$2, D$\leq$3 and D$\leq$5 respectively). Depth D=0 means the
true facts can be ``proved'' by  simple lookup in the context (no inference).
The fifth dataset, called DMax, contains questions up to depth 5, and is used
to test generalization to depths unseen in training on the other four datasets.

\subsection{Theory Generation \label{theory-generation}}

Theories contain two types of facts:
\begin{ite}
\item attributes is$(e_i,a_j)$ e.g., is(Alan,Big). 
\item relations $r_k(e_i,e_k)$ e.g., eats(Dog,Rabbit).
\end{ite}
The is$()$ predicate assigns attributes to entities, while
the $r_k()$ predicates relate two entities. Like people names,
the symbols Dog, Rabbit, etc. also denote specific entities, i.e.,
denote ``the dog'', ``the rabbit'', etc. Rules are of
the form:
\begin{quote}
{\it condition} [$\wedge$ {\it condition}]* $\rightarrow$ {\it conclusion}.
\end{quote}
The first {\it condition} is a predicate whose first argument is
a variable,\footnote{
Or with 20\% probability, an entity, in order to include some 
fully grounded rules in the datasets.}
and second argument is an attribute or entity.
For each subsequent {\it condition} and the {\it conclusion},
they are also predicates whose first argument is either the same variable
or a previously mentioned entity, and the second argument is a new attribute or entity.
(In this way, rules are constrained to have at most one
variable. Rules are implicitly universally quantified over that variable).
For example, the formal form of the first rule in Figure~\ref{example} looks:
\begin{quote}
{\it // If someone is young and round then they are kind.} \\
is(?X,Young) $\wedge$ is(?X,Round) $\rightarrow$ is(?X,Kind).
\end{quote}

Each theory contains 1-16 facts and 1-9 rules generated at random. We generate two types of theory:
\begin{enu}
\item[1.] Type 1 uses only the $is()$ predicate, with
4 entities \{Alan,Bob,...\} and 7 (non-mutually-exclusive) attributes
\{Blue,Rough,Young,...\}, drawn randomly from pools of 10 names and 14 attributes respectively.
\item[2.] Type 2 uses $is()$ and 3 other predicates \{$likes()$, $chases(), ...$\},
4 entities \{Cat,Dog,BaldEagle,...\}, and 5 attributes \{Big,Furry,...\}, drawn randomly from pools of size 6, 10, and 10 respectively.
\end{enu}
We also generate a version of each that adds negation (not)
in the facts and rule conditions/conclusions. Figure~\ref{example} is
an example of Type 1, without negation.
Figure~\ref{example2} is an example of Type 2, with negation.
Each dataset contains 100k examples (25k of each Type $\times$ without/with negation).
Data is randomly split 70/10/20 into train/dev/test partitions,
ensuring no overlap of theories between each partition. More details
of the generation procedure is described in Appendix~A.
 
% FOR THE arXiv PAPER
% 294-3 in prop-neg-depth2 modified from ``A cat likes a cat'' to ``A rabbit likes a cat''
\begin{figure}[t]
\centerline{
 \fbox{%
   \parbox{1\columnwidth}{
     {\small
The bald eagle does not eat the dog. The cat chases the dog. \\
The cat eats the bald eagle. The cat is nice. The cat likes the dog. \\
The cat likes the rabbit. The dog is furry. \\
The rabbit chases the bald eagle. The rabbit eats the bald eagle.
\vspace{1mm} \\
If someone does not eat the cat then they do not eat the dog. \\
If someone likes the bald eagle then they do not like the rabbit. \\
If someone eats the bald eagle and they do not eat the rabbit \\
\hspace*{1.5cm} then they are furry. \\
If someone is furry then they like the cat. 
\vspace{1mm} \\
% Which is true? \\
% (A) The dog does not like the cat (B) The bald eagle likes the cat \\
% (C) The rabbit likes the cat {\bf [correct]} (D) The bald eagle is furry
%Which are true? (A) The dog does not like the cat {\bf [F]} \\
%(B) The bald eagle likes the cat {\bf [F]} \\
%(C) The rabbit likes the cat {\bf [T]}
% (D) The bald eagle is furry {\bf [F]}
Q1. The bald eagle likes the cat. True/false? {\bf [F]} \\
Q2. The rabbit likes the cat. True/false? {\bf [T]} \\
Q3. The bald eagle is furry. True/false? {\bf [F]}
}}}
}   % end small
% \vspace{-1mm}
\caption{An example of a rulebase and 3 questions using relations with negation. The
reasoning for the {\bf [T]} answer is: The rabbit eats the bald eagle (given),
therefore the rabbit is furry (rule3), therefore the rabbit likes the cat (rule4).}
\label{example2}
\end{figure}

\subsection{Forward Inference \label{forward-inference}}

Given a randomly generated theory (facts+rules), we perform exhaustive forward inference
to find all its implications, noting their proof(s). (As
the domains are finite, the number of implications are
finite too). For semantics, we treat the rulebase as a logic program,
and infer the minimal, supported answer set implied by the
program \cite{Apt1988TowardsAT}. Negations in the rules' conditions are
treated as negation as failure (NAF), and we ensure that
the rulebase is stratified to avoid ambiguity and cycles \cite{Bidoit1991GeneralLD}.
Inference is performed layerwise to find the minimal supported model,
and inconsistent and unstratified rulebases are discarded.
We also check that inference proceeds to the depth
required for the target dataset, e.g., for the D$\leq$3 dataset,
at last one fact must require depth 3 inference to infer it for
all its theories.
% In addition, 
% all theories must have at least one inference at depth 1
% so as not to be trivial. (For the D=0 and D$\leq$1 datasets,
% we only test with conclusions found at depths 0 and $\leq$1).

\subsection{Question Generation and English Synthesis \label{question-generation}}

For each theory, we generate several questions with answer `true'
by selecting from the inferred facts, 
one at each depth of inference from 0 to the dataset's target
depth (e.g., for the D$\leq$2 dataset, we generate 3 `true' questions
at depths $d$ = 0, 1, and 2 for each theory). For each `true'
question we also generate a `false' question
by negating a conclusion proven at the same depth.
We then generate the same number of questions using facts that are unproven
(false under a closed-world assumption), drawing equally
from unproven, instantiated positive rule conclusions
or other unproven positive facts. Half are used as
questions labeled as false (via the CWA), and for
diversity, half are flipped by negating the fact and changing
the label to true (i.e., ``$f$? False'' becomes ``Not $f$? True'').
Thus a theory for depth $d$ has (up to) 4(d+1) questions, with an
equal balance of true and false answers. Each
question is also annotated with the inference
depth needed to answer it.\footnote{For proven facts, we use the depth of the shallowest proof tree.
For unproven facts, we use the depth of the shallowest branch of the proof tree that fails
(or the deepest of the shallowest if there are multiple (failed) proofs).}

Finally the theories and questions are converted into (synthetic)
English, using simple natural language templates plus rules to
improve fluency (e.g., using pronouns). We use three templates
(randomly selected per rule): ``If {\it condition} {\it [}and {\it condition]*} then {\it conclusion}.'',
``All {\it attribute*} people$\vert$things are {\it attribute}.'',
and ``{\it attribute*} people$\vert$things are {\it attribute}.'',
the last two only applicable to rules involving just attributes.
Examples are shown in Figures~\ref{example} and~\ref{example2}.   % FOR THE arXiv

\section{Experiments}

% Our goal is to establish whether a transformer can reason with rules --
%or at least reliably emulate the i/o behavior of a system that can
%do so. Whether the transformer actually {\it is} doing
%some form of inference propagation within its layers is an open
%question, that we explore further in Section~\ref{leave-one-out}.
%In either case, the ability to derive correct conclusions from
%rules provided at runtime would be remarkable. \red{``remarkable'' - or better word?}.

We now describe five sets of experiments we have conducted, addressing
the five questions described in the Introduction.

\subsection{Models}

We conduct all our experiments (bar Section~\ref{other-architectures-section}) 
using RoBERTa-large, % \cite{roberta} (an extensively pretrained variant of BERT),
additionally fine-tuned on the RACE dataset \cite{race}. This additional fine-tuning step
has been previously shown to help with sensitivity to hyperparameters \cite{Phang2018Stilts} and
improve question-answering \cite{Sun2018ImprovingMR}. We use fixed hyperparameters (learning rate etc),
inheriting the settings from RoBERTa on RACE \cite{roberta}.  

We train RoBERTa to predict true/false (i.e., binary classification)
for each question statement. Questions are supplied to RoBERTa
as: {\it [CLS] context [SEP] statement [SEP]}, where {\it context} is the theory (facts+rules,
expressed in language) and {\it statement} is the fact to try and prove. % (or label false, under a CWA).
The {\it [CLS]} output token is projected to a single logit. 
A logit score of $>$0 is treated as predicting true, otherwise the answer is false.
Training is performed using cross-entropy loss.
For evaluation, we measure accuracy. (The test data has an equally balance of TRUE/FALSE answers,
hence the baseline of random guessing is 50\%).

\subsection{Can RoBERTa Answer Reasoning Questions?}
\label{main}

% \subsubsection{Main Evaluation}

\begin{table}
\centering
% PPTX version of the table
\includegraphics[width=1\columnwidth]{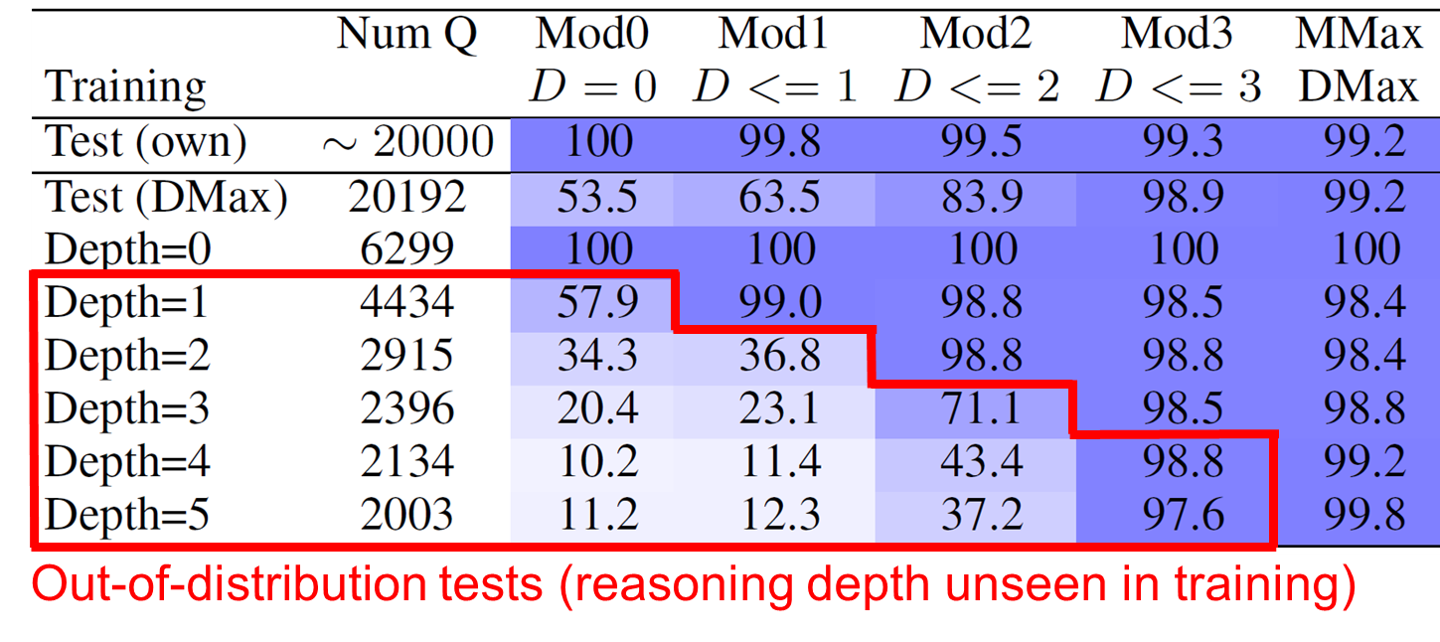}
\eat{  % REPLACED WITH THE PPTX-ANNOTED VERSION OF THIS TABLE
{\small
\setlength{\tabcolsep}{3pt}	% narrower columns
\begin{tabular}{lcccccc}
\hline
 & Num Q & Mod0 & Mod1 & Mod2 & Mod3 & \MMax\\
Training &  & $D=0$ & $D<=1$ & $D<=2$ & $D<=3$ & \DMax\\\hline
Test (own) & $\sim20000$ & \cellcolor{blue!50} 100 & \cellcolor{blue!50} 99.8 & \cellcolor{blue!50} 99.5 & \cellcolor{blue!50} 99.3 & \cellcolor{blue!50} 99.2\\\hline
Test (\DMax) & 20192 & \cellcolor{blue!27} 53.5 & \cellcolor{blue!32} 63.5 & \cellcolor{blue!42} 83.9 & \cellcolor{blue!49} 98.9 & \cellcolor{blue!50} 99.2\\
Depth=0 & 6299 & \cellcolor{blue!50} 100 & \cellcolor{blue!50} 100 & \cellcolor{blue!50} 100 & \cellcolor{blue!50} 100 & \cellcolor{blue!50} 100\\
Depth=1 & 4434 & \cellcolor{blue!29} 57.9 & \cellcolor{blue!50} 99.0 & \cellcolor{blue!49} 98.8 & \cellcolor{blue!49} 98.5 & \cellcolor{blue!49} 98.4\\
Depth=2 & 2915 & \cellcolor{blue!17} 34.3 & \cellcolor{blue!18} 36.8 & \cellcolor{blue!49} 98.8 & \cellcolor{blue!49} 98.8 & \cellcolor{blue!49} 98.4\\
Depth=3 & 2396 & \cellcolor{blue!10} 20.4 & \cellcolor{blue!12} 23.1 & \cellcolor{blue!36} 71.1 & \cellcolor{blue!49} 98.5 & \cellcolor{blue!49} 98.8\\
Depth=4 & 2134 & \cellcolor{blue!5} 10.2 & \cellcolor{blue!6} 11.4 & \cellcolor{blue!22} 43.4 & \cellcolor{blue!49} 98.8 & \cellcolor{blue!50} 99.2\\
Depth=5 & 2003 & \cellcolor{blue!6} 11.2 & \cellcolor{blue!6} 12.3 & \cellcolor{blue!19} 37.2 & \cellcolor{blue!49} 97.6 & \cellcolor{blue!50} 99.8\\
\hline
\end{tabular}
} % end small
} % end eat
\vspace{-4mm}
\caption{Accuracy of models (Mod0,...) trained and tested on the five datasets
(``Test (own)'' row), and tested on all, and different slices, of the \DMax~test set.
The boxed area indicates test problems at depths unseen during training.}
% (The random baseline for these and all other results is 50\%, as the datasets are balanced true/false).
\label{main-results}
\end{table}

\begin{figure}
\centering
    \includegraphics[width=0.8\columnwidth]{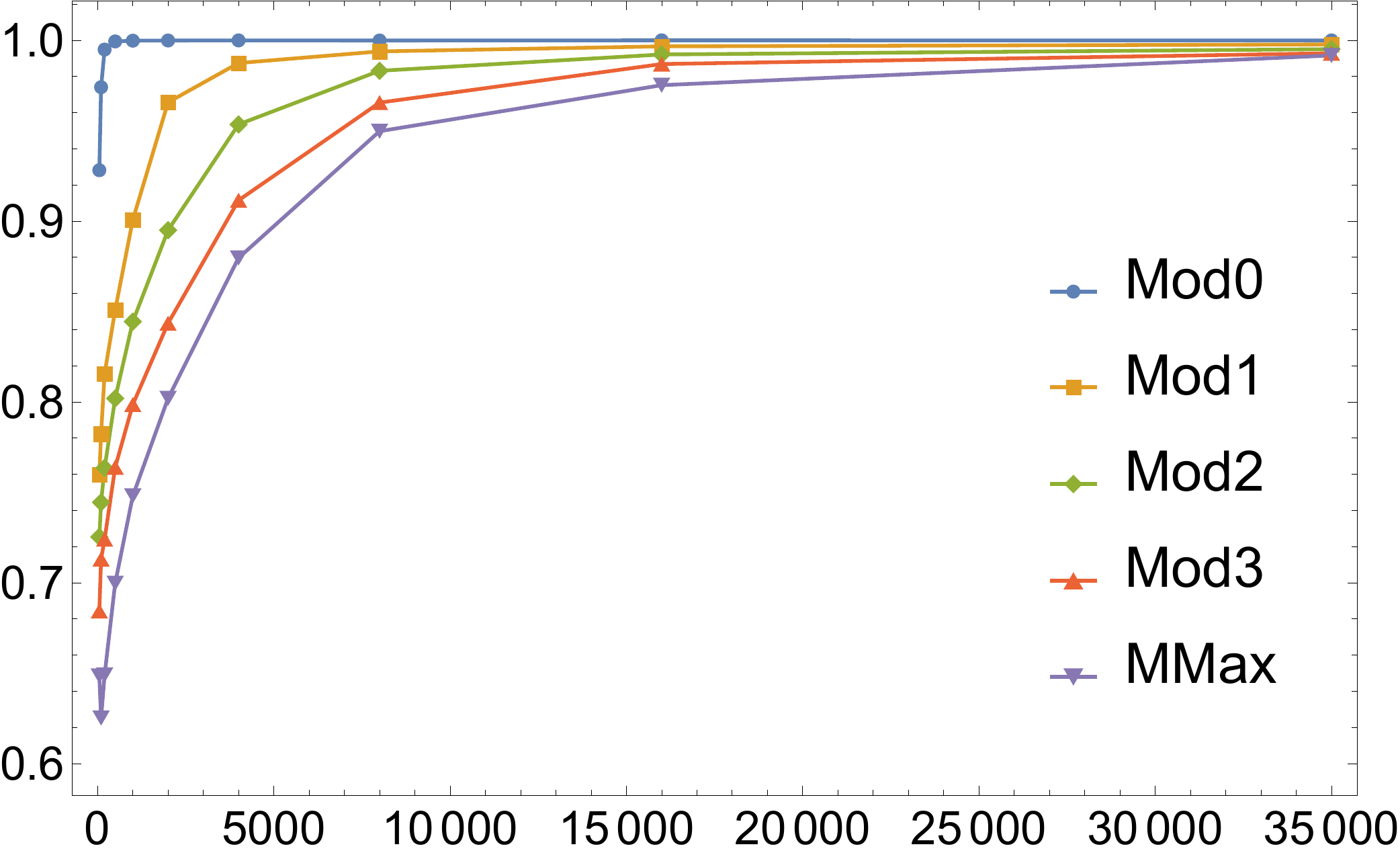}
    \vspace{-2mm}
    \caption{Learning rates (in-distribution test accuracy vs. number of training examples).
    The lower depth models require less training data, but do not generalize well (boxed area in Table~\ref{main-results}).}
    \label{learning-curves}
\end{figure}

We train and test RoBERTa models on each of our datasets D=0, D$\leq$1, D$\leq$2, D$\leq$3, and \DMax,
containing problems requiring reasoning up to depths 0, 1, 2, 3, and 5 respectively.
% We call the resulting models Model0, Model1, Model2, and Model3, as we will
% refer to them later.
We then test the models on the \DMax~dataset, that includes problems at
depths greater than the other datasets. The results are shown in Table~\ref{main-results}.
The results suggest the following findings:
\begin{enu}
\item[1.] RoBERTa is able to {\bf master the test data almost perfectly} (99\% accuracy, row 1)
even though the specific reasoning problems (facts+rules) in each test question
are distinct from those in the training set.
\item[2.] The Depth=0 model, Mod0, only trained on lookup questions, is (unsurprisingly)
{\bf unable to answer questions requiring reasoning} (column Mod0).\footnote{ \label{below-random-footnote}
In fact, we see an interesting learning artifact, namely Mod0 scores worse
than random (50\%) at depths higher than 2.
This arises because most questions at these depths are provably true facts,
but Mod0 learns to predict all facts are false except those explicitly
given (as that is all it has seen at training time), hence
systematically gets these wrong.}
% In fact, we see an interesting learning artifact, namely Mod0 scores worse than
% random (50\%) at depths higher than 2. This arises as Mod0 learns to 
% predict all (positive) facts are false except the explicitly given facts,
% as that is all it has seen at training time. 
% However, most (90\%) of the facts proven at depth 2+
% in \DMax~are positive facts proven true, hence Mod0 gets these all wrong.
% In a few cases, a fact is proven false at depth 2+, and thus Mod0 will
% by chance get these right. However, it is rare (10\%) to see a
% negated fact with a proof, as half the theories do not use negation, and
% for the remainder, negative rule conclusions are generated with lower probability
% than positive rule conclusions during dataset generation.
% We see a similar artifact with the Mod1 model.}
\item[3.] As we train with increasingly deep inference, the models' ability
to generalize improves. The D$\leq$2 model (questions involving problems
up to depth 2) achieves 71.1\% on Depth=3 problems, while
{\bf the D$\leq$3 model generalizes well} right up to the maximum depth tested (e..g, 97.6\% for Depth=5 problems).
% but little more than random at
% higher depths. The D$\leq$2 model starts to perform well at higher depths,
% even scoring 68\% F1 on depth 7 problems. {\bf The D$\leq$3 model is largely able to solve (score 85\%+) problems at all depths tested}.
\end{enu}
\noindent
Looking at the models' accuracies (on their own in-distribution test data) as a function of training size,
we see the higher depth models require more data to learn (Figure~\ref{learning-curves}), but have better
out-of-domain performance (boxed area in Table~\ref{main-results}).

% \red{Could mention train on D=3, D$\leq$3small (superceded by learning curves) and train on D<=5?}

% \subsubsection{Robustness Testing \label{robustness}}

We additionally test the robustness of the models' answers
by perturbing the original theories. Specifically, for each test
fact $f$ that is true, we test whether removing a sentence that is part
of the proof of $f$ causes the prediction to (desirably) flip from true to false.
We call these sentences in the proof tree {\it critical sentences}, as
the truth of $f$ depends on them. Conversely, removing an
{\it irrelevant} sentence should cause no change to the model's
prediction. As we know the original proof trees for each fact $f$
in the dataset, we can identify the critical and irrelevant
sentences by simple inspection of those trees.\footnote{
If there are multiple, alternative proofs for $f$, we define
a critical sentence as one that is used in {\it all} the proofs.
To support this, we generate and record all possible proofs
for each provable fact $f$ during dataset generation.} Typically,
1-6 sentences of the $\approx$15-20 sentences are critical for
proving each provable fact. % (Figure~\ref{critical-sentence-counts}).

\eat{
% OR MOVE LATER
\begin{figure}
\centering
\includegraphics[width=1\columnwidth]{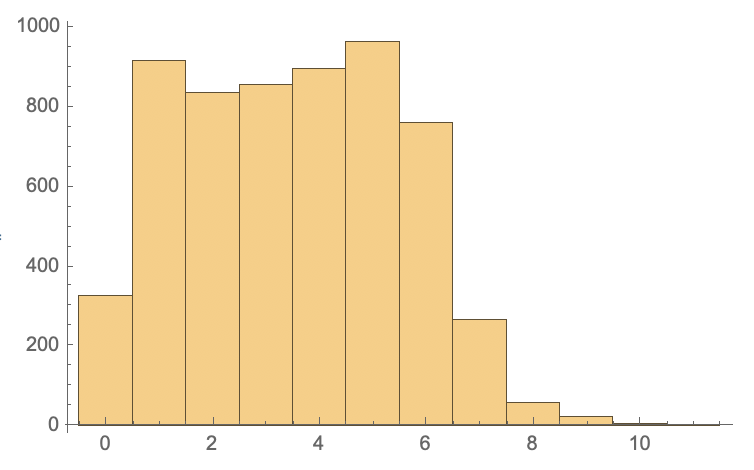}
% \vspace{-1mm}
\caption{Counts of the number of sentences critical for proving each proven fact $f$ in the DMax test set (no negation subset).
Typically 1-6 sentences (from theories of size $\approx$15-20 sentences) are critical.}
\label{critical-sentence-counts}
\end{figure}
}

We test this using the no-negation\footnote{
With negation, the definition of critical sentence becomes more
complex because the the theory is non-monotonic (i.e., {\it removing} a sentence
may cause a fact to become {\it true}). Hence, we omit theories with negation
for this analysis.} half of the \DMax~test set ($\approx$10k questions).
In this partition, 5904 questions have
proofs (are true). (The remaining % 4116 of them
questions are false under the CWA).\footnote{
For questions that were flipped from ``$f$? False'' to ``Not $f$? True''
to encourage diversity, we undo the flips and just consider
the positive forms of facts $f$.}
% --- end footnote
For each of these questions,
we remove each of the theory sentences $s_i$ in turn, and measure the prediction accuracy
on each result. As there are about 19 sentences/theory on average, this results in
113978 ``sentence removed'' probes (of which 20746 have a critical sentence removed,
and 93232 have an irrelevant sentence removed). Ideally, removing a sentence
critical to a question $f$ should flip the model's prediction from T to F, while removing a noncritical
sentence should leave the prediction unchanged as T. We also measure
overall performance on the entire dataset of questions with perturbed theories.

The results are shown in Tables~\ref{robustness1} and~\ref{robustness2}.
We observe:
\begin{enu}
\item[1.] The {\bf overall accuracy is largely unchanged} on the full collection of
questions with perturbed theories, suggesting robustness to these variants
(last column, Table~\ref{robustness1}).
\item[2.] For the (20k) questions where the prediction is expected to flip from
true to false, we see this flip occurs 81\% of the time, Table~\ref{robustness2}.
This suggests {\bf moderate robustness} to this specific type of perturbation, although
notably less than for a formal theorem prover (that would make this flip 100\% of the time).
% (and rarely the other way from false to true)
For the remaining (93k) questions, the prediction (correctly) stays true over 99\% of the time (no Table).
% \footnote{
% \label{noncritical-data-footnote}
% The analogous data for removing a noncritical sentence is:
% \green{169780} T$\rightarrow$T, \red{329} T$\rightarrow$F, \green{720} F$\rightarrow$T, \red{1481} F$\rightarrow$F.}).

\end{enu}

% NEW	T	169780	720
% 	F	329	1481

% NEW	T	169780	720
% 	F	329	1481

\begin{table}
\centering
{\small
\begin{tabular}{lcccc}
\hline
 & Original & Remove & Remove & Remove\\
 &  & Irrelevant & Critical & Any\\
\hline
% OLD FIGURES
% Accuracy (test) & \cellcolor{blue!43} 93.2 & \cellcolor{blue!44} 94.4 & \cellcolor{blue!38} 88.3 & \cellcolor{blue!43} 93.3\\
% NEW FIGURES
  Accuracy (test) & \cellcolor{blue!43} 99.4 & \cellcolor{blue!44} 99.6 & \cellcolor{blue!38} 81.2 & \cellcolor{blue!43} 96.3\\
\hline
\end{tabular}
}
% \vspace{-1mm}
\caption{Accuracy on the DMax (no negation) subset, and all its (113k) perturbed (one context sentence removed) variants. The overall accuracy
(Remove Any, last column) is largely unchanged, but with a drop for the subset where a critical sentence was removed.}
\label{robustness1}
\end{table}

\begin{table}
\centering
{\small
    \setlength{\tabcolsep}{3pt}	% narrower columns
\begin{tabular}{|lr|ll|} \hline
 &  & \multicolumn{2}{|c|}{\bf Original predictions for true (positive) facts:} \\
 &  &  {\bf T} & {\bf F} \\ \hline
 {\bf New} & {\bf T}    & {\bf \red{3895}} (should have flipped)    & {\bf \red{10}} (incorrectly flips)        \\
 {\bf Pred.} & {\bf F}  & {\bf \green{16654}} (correct flips)       &  {\bf \green{187}} (becomes correct) \\ \hline
\end{tabular}
% TOTAL: 20746 questions
}
% \vspace{-1mm}
% \caption{Removing a sentence critical to proving a true (positive) fact $f$ correctly changes the predicted answer from T to F
% over 80\% (= 16654/(16654+3895)) of the time (column 1), demonstrating appropriate sensitivity to small theory changes. In a few (197) cases, the predicted answer was incorrect to start with (column 2). When an {\it irrelevant} sentence is removed, the predicted answer stays correct (T) over 99\% of the time (not shown in this table).}
\caption{On the true questions that were originally answered correctly (column 1),
the predicted T answer should flip to predicted F when a critical sentence is
removed. In practice, we observe this happens 81\% of the time (16654/(16654+3895)).
In a few (197) cases, the predicted answer was incorrect to start with (column 2).
When an {\it irrelevant} sentence is removed, the predicted answer stays
% correct (T) over 99\% of the time (not shown\footnotemark[\ref{noncritical-data-footnote}]).}
% correct (T) over 99\% of the time (not shown$^{\ref{noncritical-data-footnote}$).}
% correct (T) over 99\% of the time (not shown\textsuperscript{\ref{noncritical-data-footnote}}).}
correct (T) over 99\% of the time (not shown).}
\label{robustness2}
\end{table}
% TABLE FOR non-critical probes (both proved T, and 
% 		ORIGINAL PREDICTION
%		T	F
% NEW	T	169780	720
% 	F	329	1481
% OVERALL: Flips T to F = 16654+329 / 113978
\subsection{Performance on Hand-Authored Problems \label{hand-authored}}

To further test robustness and out-of-distribution performance, we test the trained
models on two hand-authored reasoning problems, both including reasoning with negation,
written independently of our datasets. Note that these new datasets are used
purely as test sets (no training on them, i.e., zero-shot performance); 
their vocabulary of entities, attributes, and predicates (bar {\it is()})
are all new to the models at test time. The two test datasets are as follows:

% \subsubsection{Test Datasets}

\paragraph{Birds}

The ``birds'' rulebase is a well-known logic problem illustrating the use of ``abnormality'' predicates \cite{McCarthy1984ApplicationsOC}.
We entered Sergot's formulation of it\footnote{https://www.doc.ic.ac.uk/$\sim$mjs/teaching/KnowledgeRep491/ ExtendedLP\_491-2x1.pdf, p5,
also in Appendix~B.} verbatim (bar syntax), and generated a series of test questions using the same procedure as earlier.
Figure~\ref{birds} illustrates the problem (in restricted English, exactly as presented to our model) and four example questions.
We created two linguistic expressions of the formal theory, Birds1 and Birds2. Birds2 is shown in Figure~\ref{birds}, while Birds1 is
identical except ``can/cannot fly'' is replaced with ``is/is not flying'' to make the negation (``not'') more explicit (this turns out
not to matter).

\begin{figure}[t]
\centerline{
 \fbox{%
   \parbox{1\columnwidth}{
     {\small
If someone is a bird and not abnormal then they can fly. \\    
If someone is an ostrich then they are a bird. \\
If someone is an ostrich then they are abnormal. \\
If someone is an ostrich then they cannot fly. \\
If someone is a bird and wounded then they are abnormal. \\
If someone is wounded then they cannot fly.
\vspace{1mm} \\
Arthur is a bird. Arthur is not wounded. Bill is an ostrich. \\
Colin is a bird. Colin is wounded. \\
Dave is not an ostrich. Dave is wounded. 
\vspace{2mm} \\
% Which is true? A) Arthur can fly {\bf [correct]} (B) Bill can fly \\
% (C) Colin can fly (D) Dave can fly
% Which are true? (A) Arthur can fly {\bf [T]} (B) Bill can fly {\bf [F]} \\
%(C) Colin can fly {\bf [F]} (D) Dave can fly {\bf [F]}
\eat{Q1. Arthur can fly. True/false? {\bf [Answer: T]} \\
Q2. Bill can fly. True/false? {\bf [F]} \\
Q3. Colin can fly. True/false? {\bf [F]} \\
Q4. Dave can fly. True/false? {\bf [F]}
}
\setlength{\tabcolsep}{3pt}	% narrower columns
\begin{tabular}{ll}
\hspace*{-2.5mm} Q1.Arthur can fly. True/false? {\bf [T]} & Q2.Bill can fly. True/false? {\bf [F]} \\
\hspace*{-2.5mm} Q3.Colin can fly. True/false? {\bf [F]} & Q4.Dave can fly. True/false? {\bf [F]}
\end{tabular}
}}}
}   % end small
% \vspace{-1mm}
\caption{Sergot's ``birds'' puzzle includes reasoning about abnormality predicates.
The dataset contains these and other questions about the single theory.}
\label{birds}
\end{figure}

\paragraph{Electricity}

\begin{figure}[t]
\centerline{
 \fbox{%
   \parbox{1\columnwidth}{
     {\small
\begin{tabular}{ll}
The circuit has a switch. & \hspace*{15mm} \multirow{3}{*}{\includegraphics[height=14mm]{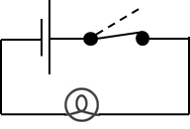}} \\
The switch is on. & \\
The circuit has a light bulb. & \\
 & \\
\end{tabular}  \\
If a circuit has a switch and the switch is on \\
\hspace*{1cm} then the circuit is complete. \\
If a circuit does not have a switch then the circuit is complete. \\
If a circuit is complete then a current runs through the circuit. \\
If a current runs through a circuit and the circuit has a light bulb \\
\hspace*{1cm} then the light bulb is glowing. \\
If a current runs through a circuit and the circuit has a bell \\
\hspace*{1cm} then the bell is ringing. \\
If a current runs through a circuit and the circuit has a radio \\
\hspace*{1mm} then the radio is playing.
\vspace{1mm} \\
% Which is true? \\
% (A) A radio is playing (B) A light bulb is glowing \\
% (C) A bell is ringing {\bf [correct]} (D) A circuit has a radio
%Which are true? \\
%\hspace*{2mm} (A) The radio is playing {\bf [F]} (B) The light bulb is glowing {\bf [T]} \\
%\hspace*{2mm} (C) The bell is ringing {\bf [F]} (D) The circuit has a radio {\bf [F]}
Q1. The circuit is not complete. True/false? {\bf [F]} \\
Q2. The light bulb is glowing. True/false? {\bf [T]} \\
Q3. The radio is playing. True/false? {\bf [F]} 
}}}
}   % end small
% \vspace{-1mm}
\caption{The simple Electricity2 rulebase, an example circuit, and 3 questions about the circuit. (Circuit diagram is for illustration only).}
\label{electricity}
\end{figure}

We also created a small rulebase about an electrical circuit, describing the conditions
for an appliance to function. We created 4 variants of increasing complexity, containing
5, 6, 11, and 12 rules respectively. For each rulebase, we generate different
scenarios (the facts) by randomly selecting from possible ground facts. Questions
are then generated against each scenario using the same procedure as
earlier, resulting in 4 test sets.
Figure~\ref{electricity} shows the Electricity2 rulebase with an example scenario plus three questions.
(Appendix~C shows all four rulebases).

\subsubsection{Results \label{hand-authored-results}}

The results are in Table~\ref{handcrafted}, tested using the earlier
trained models. Note that these new problems and vocabularies were unseen
during training (i.e., are zero-shot). We observe:
\begin{enu}
\item[1.] The ``birds'' problems are {\bf solved (almost) perfectly} by all but the
non-reasoning (Mod0) model (MMax gets one question wrong on Birds1).
\item[2.] The MMax~model (trained on \DMax) {\bf solves all but one}
of these datasets with 90\%+ scores.
\end{enu}
These are two point demonstrations that the trained models can be
used to solve novel reasoning problems with high reliability (90\%+
in all but one case).

We see one surprising anomaly also: the models trained with deeper reasoning
depths do slightly worse on Electricity4 than the depth 1 model, Mod1.
From investigation, we find almost all failing questions at higher depths are those
where the queried fact $f$ is an unsatisfied rule conclusion (hence should be false),
in particular when the first argument of $f$ is not the first argument of one
of the rule's conditions. Because of the way the original dataset was generated, examples
similar to this are very rare in the training data, possibly causing
this anomaly. More generally this illustrates that even when trained
on a diversity of problems, the trained model can have unanticipated blind-spots.
We discuss this further in Section~\ref{generating-training-data}.

\subsection{Reasoning with Paraphrased Rules}

Our experiments so far have been with synthetic language, but our
ultimate goal is to reason over full natural language.
% , e.g., with sentences drawn from a corpus.
To test transfer to more natural linguistic forms, we generated a new dataset of
40k examples, using crowdworkers to paraphrase our theories. Of course, this only
tests robustness to paraphrasing, not to abitrary natural language. Nevertheless,
it is a small first step in this direction.

To generate our data, we follow a similar approach to \cite{clutrr}, described below.
For this experiment, we used Type 1 theories % (i.e., using just the is() predicate)
without negation, i.e., the same form as in Figure~\ref{example}.

\begin{table}
\centering
\includegraphics[width=1\columnwidth]{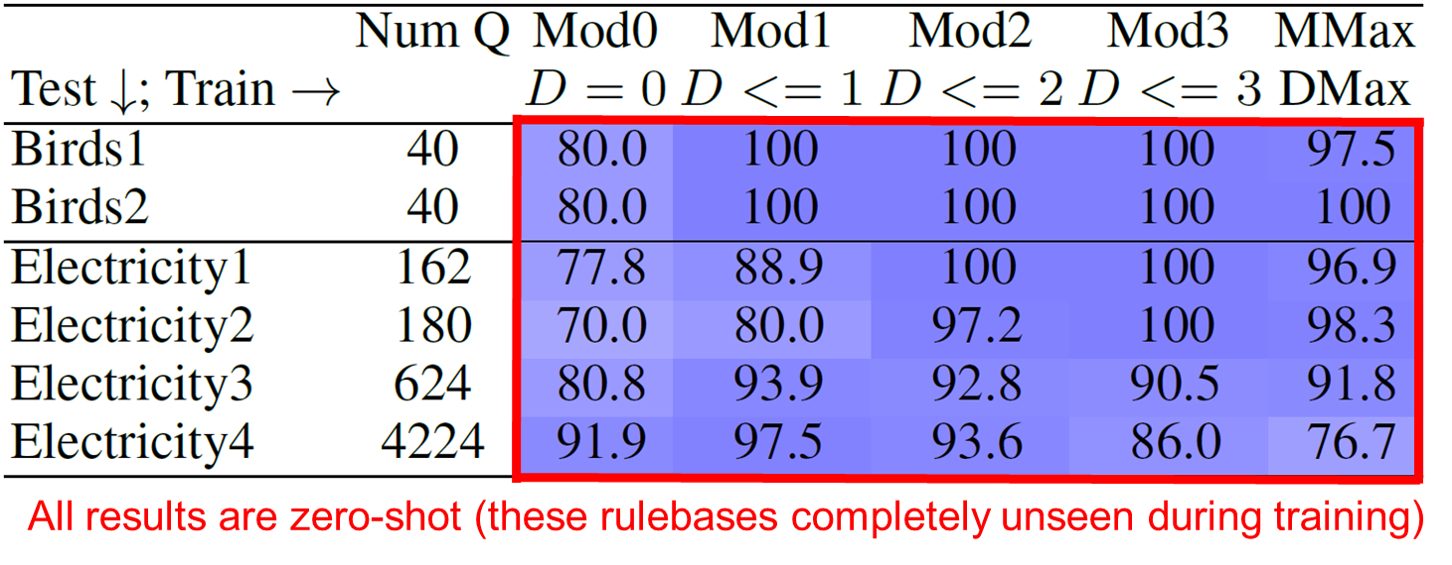}
\eat{
{\small
    \setlength{\tabcolsep}{1pt}	% narrower columns
\begin{tabular}{lcccccc}
\hline
 & Num Q & Mod0 & Mod1 & Mod2 & Mod3 & \MMax\\
Test $\downarrow$; Train $\rightarrow$ &  & $D=0$ & $D<=1$ & $D<=2$ & $D<=3$ & \DMax\\\hline
Birds1 & 40 & \cellcolor{blue!40} 80.0 & \cellcolor{blue!50} 100 & \cellcolor{blue!50} 100 & \cellcolor{blue!50} 100 & \cellcolor{blue!49} 97.5\\
Birds2 & 40 & \cellcolor{blue!40} 80.0 & \cellcolor{blue!50} 100 & \cellcolor{blue!50} 100 & \cellcolor{blue!50} 100 & \cellcolor{blue!50} 100\\\hline
Electricity1 & 162 & \cellcolor{blue!39} 77.8 & \cellcolor{blue!44} 88.9 & \cellcolor{blue!50} 100 & \cellcolor{blue!50} 100 & \cellcolor{blue!48} 96.9\\
Electricity2 & 180 & \cellcolor{blue!35} 70.0 & \cellcolor{blue!40} 80.0 & \cellcolor{blue!49} 97.2 & \cellcolor{blue!50} 100 & \cellcolor{blue!49} 98.3\\
Electricity3 & 624 & \cellcolor{blue!40} 80.8 & \cellcolor{blue!47} 93.9 & \cellcolor{blue!46} 92.8 & \cellcolor{blue!45} 90.5 & \cellcolor{blue!46} 91.8\\
Electricity4 & 4224 & \cellcolor{blue!46} 91.9 & \cellcolor{blue!49} 97.5 & \cellcolor{blue!47} 93.6 & \cellcolor{blue!43} 86.0 & \cellcolor{blue!38} 76.7\\
\hline
\end{tabular}
}
} % end eat
\vspace{-7mm}
\caption{Accuracy of the earlier models tested on hand-crafted rulebases (zero shot, no fine-tuning).
Note that the models were {\it only} trained on the earlier datasets (e.g., Figures~\ref{example} and~\ref{example2}),
and thus the new rulebases' entities, attributes, and predicates (bar {\it is()}) are completely unseen until test time.}
\label{handcrafted}
\end{table}

\subsubsection{Dataset Generation}

To generate the new dataset, called ParaRules, we first generated a novel collection
of 10k theories (facts+rules) expressed in synthetic language, as before,
then extracted the ``fact groups'' and rules from each.
A ``fact group'' is all the facts in a theory about a particular
person, e.g., (from Figure~\ref{example}) ``Alan is blue. Alan is rough. Alan is young.'',
while a rule is just the original ``If...then...'' sentence.
We then asked crowdworkers to creatively re-express the fact-groups and rules,
shown to them in English, in their own words. For example, the
earlier fact-group might be rewritten as: 
``Alan is on the young side, but rough. He often feels rather blue.''.
Rewritten fact-groups were then turned into templates by variabilizing the
person name. Turkers also rephrased each rule (no variabilization needed).
Rephrasings were automatically checked to make sure that all the key attributes
were mentioned (and no others included), and rejected otherwise.

We use these to assemble the new ParaRules dataset of 40k questions against $\approx$2k
theories expressed in the paraphrased language.
To build each theory, facts were collected by randomly
sampling and instantiating fact-group templates with people's names,
and rules were randomly sampled. An example is shown in Figure~\ref{nlp-example}.
The train, dev, and test sets were generated using different
partitions of the templates, to ensure that no templates were
shared between partitions.

As we kept track of the corresponding logic underlying each fact
group and rule, we can then generate questions as before: Exhaustively
forward-chain on the (logic version of) the theory, discard if a
contradiction is hit or reasoning is of insufficient depth (we require
at least depth 3 reasoning), and then for each depth select inferred
and non-inferred facts as true/false questions as before.

% Example Turk 1210-3
\begin{figure}[t]
\centerline{
 \fbox{%
   \parbox{1\columnwidth}{
     {\small
Alan, who is round, red, kind, and also green, tends to be rather blue. 
In the snow sits Bob, crying from being cold. 
Charlie has green teeth and rough skin. People also notice his blue eyes. 
\vspace{1mm} \\
A quite nice person who is red and green is also big. \\
Any big, kind person that turns red is cold to the touch. \\
Young, kind people have a habit of being nice. \\
A kind person will certainly be young. 
\vspace{1mm} \\
% Which is true? (A) Dave is nice (B) Charlie is big \\
% (C) Alan is nice {\bf [correct]} (D) Bob is nice
Q1. Dave is nice. True/false? {\bf [F]} \\
Q2. Charlie is big. True/false? {\bf [F]}\\
Q3. Alan is nice. True/false? {\bf [T]}
% Which are true? (A) Dave is nice {\bf [F]} \\
% (B) Charlie is big {\bf [F]} (C) Alan is nice {\bf [T]} (D) Bob is nice {\bf [F]}
}}}
}   % end small
% \vspace{-1mm}
\caption{A paraphrased theory in the ParaRules dataset.
The reasoning for the true answer here is: Alan is kind (given), therefore
young (rule4), therefore nice (rule3).
\label{nlp-example}}
\end{figure}

\begin{table}
\centering
% PPT-edited version
% \includegraphics[width=1\columnwidth]{natlang-results.png}
\includegraphics[width=1\columnwidth]{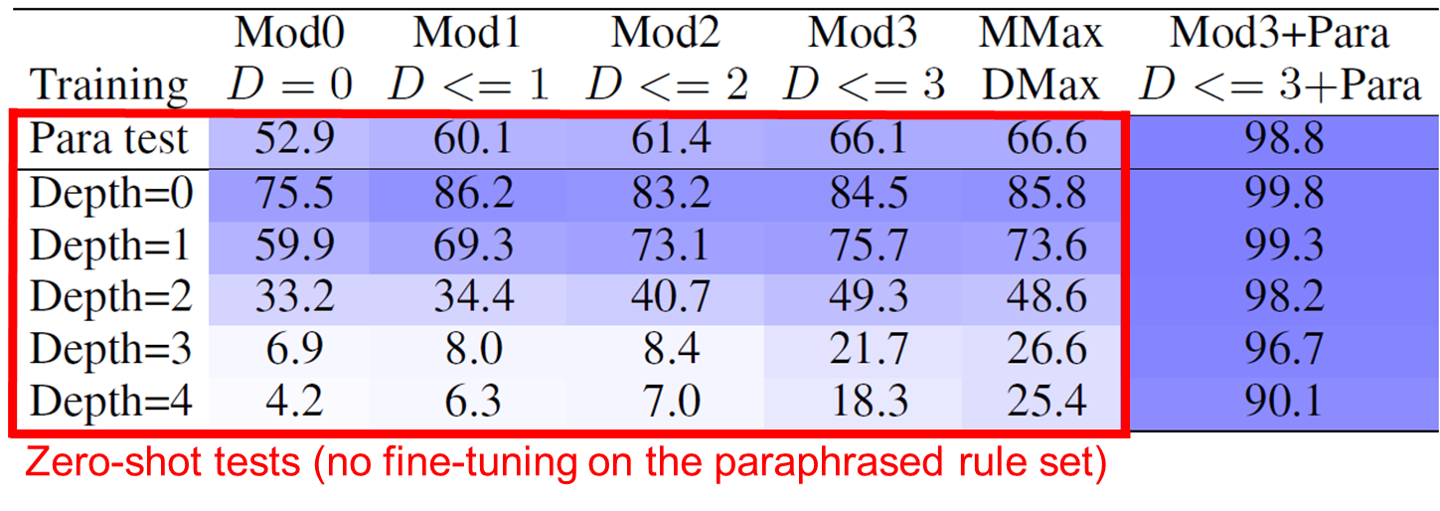}
\eat{
{\small
    \setlength{\tabcolsep}{3pt}	% narrower columns
\begin{tabular}{lcccccc}
\hline
 & Mod0 & Mod1 & Mod2 & Mod3 & \MMax & Mod3+Para\\
Training & $D=0$ & $D<=1$ & $D<=2$ & $D<=3$ & \DMax & $D<=3+$Para\\\hline
Para test & \cellcolor{blue!26} 52.9 & \cellcolor{blue!30} 60.1 & \cellcolor{blue!31} 61.4 & \cellcolor{blue!33} 66.1 & \cellcolor{blue!33} 66.6 & \cellcolor{blue!49} 98.8\\\hline
Depth=0 & \cellcolor{blue!38} 75.5 & \cellcolor{blue!43} 86.2 & \cellcolor{blue!42} 83.2 & \cellcolor{blue!42} 84.5 & \cellcolor{blue!43} 85.8 & \cellcolor{blue!50} 99.8\\
Depth=1 & \cellcolor{blue!30} 59.9 & \cellcolor{blue!35} 69.3 & \cellcolor{blue!37} 73.1 & \cellcolor{blue!38} 75.7 & \cellcolor{blue!37} 73.6 & \cellcolor{blue!50} 99.3\\
Depth=2 & \cellcolor{blue!17} 33.2 & \cellcolor{blue!17} 34.4 & \cellcolor{blue!20} 40.7 & \cellcolor{blue!25} 49.3 & \cellcolor{blue!24} 48.6 & \cellcolor{blue!49} 98.2\\
Depth=3 & \cellcolor{blue!3} 6.9 & \cellcolor{blue!4} 8.0 & \cellcolor{blue!4} 8.4 & \cellcolor{blue!11} 21.7 & \cellcolor{blue!13} 26.6 & \cellcolor{blue!48} 96.7\\
Depth=4 & \cellcolor{blue!2} 4.2 & \cellcolor{blue!3} 6.3 & \cellcolor{blue!4} 7.0 & \cellcolor{blue!9} 18.3 & \cellcolor{blue!13} 25.4 & \cellcolor{blue!45} 90.1\\
\hline
\end{tabular}
}
} % end eat
\vspace{-7mm}
\caption{Accuracy with rules paraphrased into more natural language (ParaRules),
without fine-tuning (zero shot) and with (last column only). The strongest zero-shot model (\MMax) partially solves (66.6\%) this
problem zero-shot, with strongest performance for depth 0 and 1 inferences.}
\label{natlang}
\end{table}

\subsubsection{Results}

We ran the earlier trained models on the ParaRules test partition (no fine-tuning,
i.e., zero shot). The results are shown in Table~\ref{natlang}. The strongest
model, \MMax, partially solves this dataset with a score of 66.6\%,
higher for questions requiring less inference, and lower for questions
requiring more inference. (The below-random scores for D=0 reflect
the same artifact as earlier, namely predicting everything
as false except for facts explicitly given. See Footnote~\ref{below-random-footnote}).

Note that these results are for zero-shot, with no model exposure
to the paraphrased data during training. In contrast, we also trained a model
using {\it both} of the D$\leq$3 and ParaRules training partitions.
The resulting model (last column in Table~\ref{natlang}) has an accuracy of 98.8\% on the 
ParaRules test partition % (96.59\% on the ParaRules test, 99.21\% on D$\leq$3 test)
(even though the ParaRules test rewordings are
distinct from train and dev), showing near-perfect performance is
learnable.
Although a limited study, this suggests that % in a constrained way that
our findings may extend to rulebases expressed in more natural language.

\begin{figure*}[t]
{% Include a tight 1pt thick rule around the image:
\setlength{\fboxsep}{0pt}%
\setlength{\fboxrule}{1pt}%
% \fbox{\includegraphics[width=1\textwidth]{critical-sentences/perfect-critical-sentences-hires-noproof.png}}
\fbox{\includegraphics[width=1\textwidth]{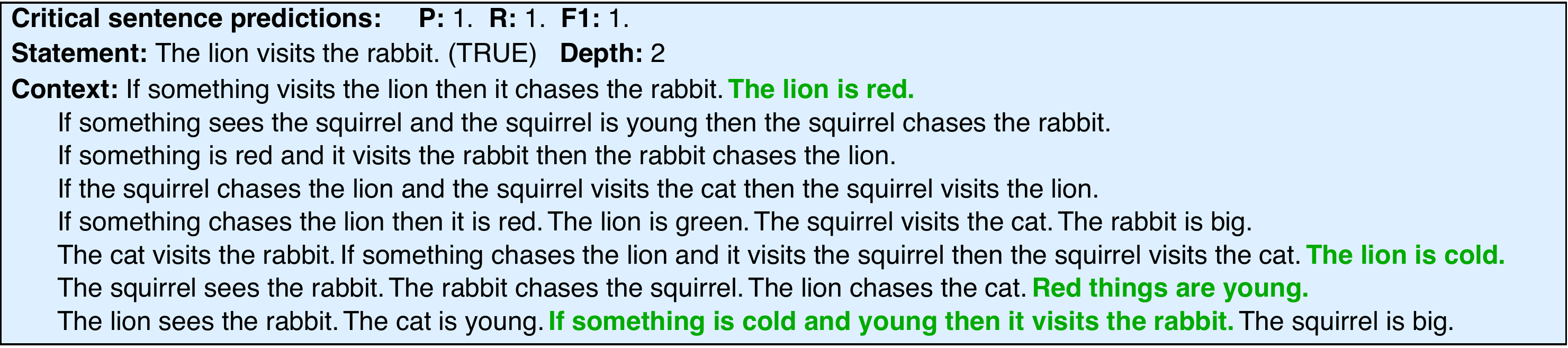}}
}%

\noindent
\small{Above, the model has correctly identified the sentences critical to the answer (shown in green).
Here, the underlying line of reasoning is:
% ``A dog likes a bald eagle (fact 8), therefore it is furry (rule 2), therefore it chases the bald eagle (rule 12).''.}
``The lion is red (first highlighted statement), therefore young (third highlighted), and as it also cold (second) it visits the rabbit (fourth).''}

\vspace{2mm}

{% Include a tight 1pt thick rule around the image:
\setlength{\fboxsep}{0pt}%
\setlength{\fboxrule}{1pt}%
% Seem to have lost the high-res version of this....
% \fbox{\includegraphics[width=1\textwidth]{critical-sentences/imperfect-critical-sentences-hires-noproof.png}}
% \fbox{\includegraphics[width=1\textwidth]{critical-sentences/imperfect-critical-sentences-noproof.png}}
\fbox{\includegraphics[width=1\textwidth]{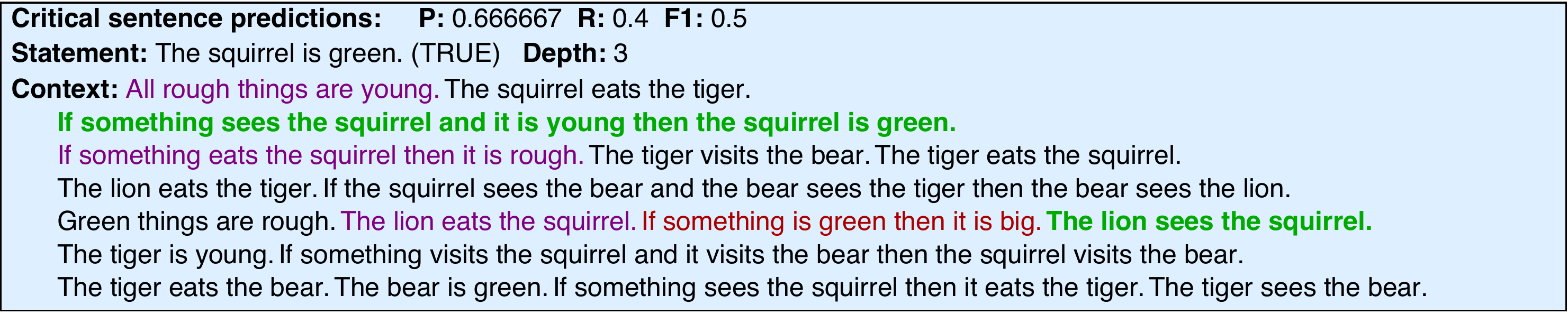}}
}%

\noindent
% for SentencePrediction5
% \small{Above, the model has identified three critical sentences (green), but has missed one (purple) and included an irrelevant fact (red).
% Here, the underlying line of reasoning is:
% ``The bald eagle chases the lion (second highlighted statement), ...
%
\small{Above, the model has identified two critical sentences (green), but has missed three (purple) and included an irrelevant rule (red).
Here, the underlying line of reasoning is:
``The lion eats the squirrel (fourth highlighted statement), therefore is rough (third), therefore young (first), and as it also sees the squirrel (sixth) it is green (second).''}
% ``A bald eagle chases a cat (fact 1), therefore it is young (rule 5), therefore furry (rule 1), and as it also chases a cat (fact 2), it is nice (rule3).''.}
\vspace{-2mm}
\begin{center}
-------------------------
\end{center}
\vspace{-2mm}
% \vspace{-3mm}
\caption{Examples of the model identifying sentences critical to the answer, perfectly (upper figure) and imperfectly (lower figure).
Perfect identification occurs for over 70\% of the provable answers (See Figure~\ref{critical-histogram} for a full histogram).}
\label{critical-sentences}
\end{figure*}

\begin{figure}
\centering
\includegraphics[width=0.7\columnwidth]{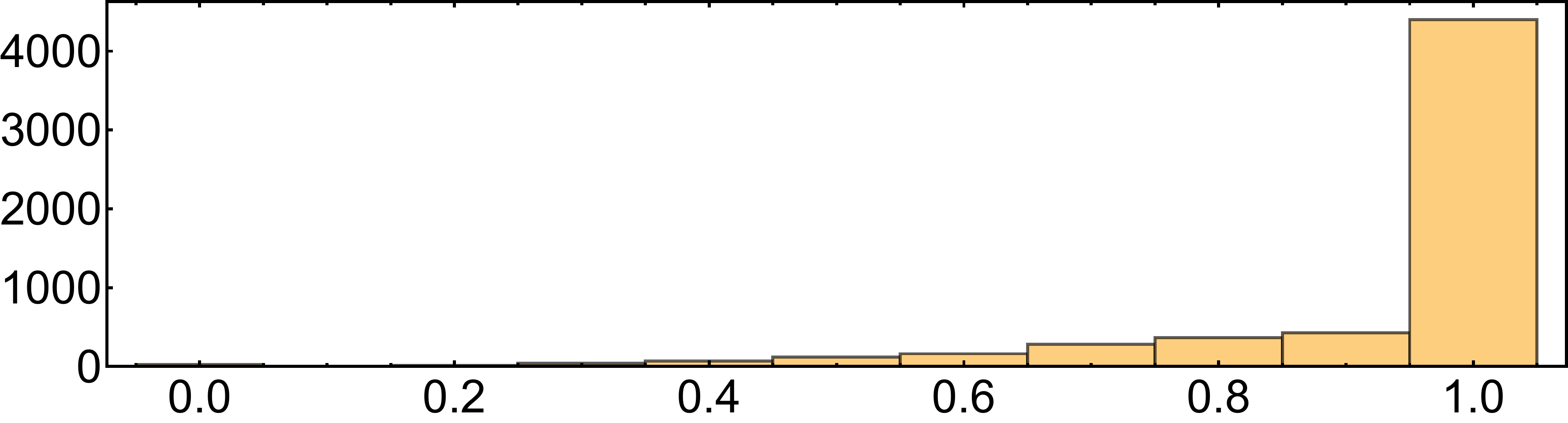}
% \vspace{-1mm}
\caption{Counts of the F1 scores for predicting which sentences are critical to the proofs of questions in DMax (test, no negation subset).
For over 70\% of the questions, the model predicts critical sentences perfectly (F1=1.0), with high F1 in the remaining case.}
\label{critical-histogram}
\end{figure}

\subsection{Generating Explanations \label{explanation}}

In Section~\ref{main}, we tested (for the no-negation theories)
whether removing a theory sentence $s_i$ caused the prediction
for a true fact $f$ to flip to false, and found that sentences
causing a flip were very often (98\%) part of the original
proof of $f$ (i.e., critical sentences), while sentences that
did not were not (97\%). Using that data about which removed sentences
caused a flip, % \footnote{We find we get slightly better results }
we can build a map of the theory paragraph showing
which sentences the model considers critical to a conclusion,
a potentially first step to providing an explanation for
the model's answers. Figure~\ref{critical-sentences} shows two examples,
the first showing where the model has perfectly identified the critical sentences,
and the second where it has made some errors.

We can quantify this ``explanatory'' performance by measuring the per-proof
scores of predicted vs. actual critical sentences for each question,
measuring the precision, recall, and F1 scores for each question in turn.
The (macro)average P/R/F1 scores are
% P=98.5, R=90.1, and F1=94.1,	% with the ``3'' heuristic
P=98.7, R=86.9, and F1=92.4, suggesting a high degree of reliability in predicting sentences critical to a proof. (This is essentially
an alternative view on the earlier robustness data, viewed from a per-proof perspective).
A histogram of the F1 scores is shown in Figure~\ref{critical-histogram}, indicating perfect critical sentence
identification for over 70\% of the questions, and high F1 for the remaining questions.
This suggests the model has some knowledge of the dependencies between the
context sentences and a particular conclusion.

\subsection{Other Architectures \label{other-architectures-section}}

To what extent are our results specific to RoBERTa? To explore this,
we also trained BERT and ESIM (an LSTM-based model for natural language inference) \cite{chen2016enhanced}
on our datasets. Also, as a sanity check that our datasets are not trivially solvable,
we ran the decomposable attention model (DECOMP) on our data \cite{Parikh2016ADA}.
% IJCAI - no table 
% Trained and tested on the \DMax~dataset,
% RoBERTa, BERT, and ESIM score an F1 of 96.0 (Table~\ref{main-results}), 86.9,
% and 29.5 respectively. Similarly, trained on the D$\leq$3 dataset and tested on \DMax
% (requiring generalization to greater inference depths), scores are 89.6 (Table~\ref{main-results}), 86.8, and 33.7 respectively.
% This indicates that BERT is also able to learn these tasks,
% i.e., our results are not specific to RoBERTa, but that the LSTM-based model ESIM
% does not perform well on these datasets, indicating both the difficulty of
% these datasets, and the power of transformers.
The results are shown in Table~\ref{other-architectures}.

\begin{table}
\centering
\includegraphics[width=1\columnwidth]{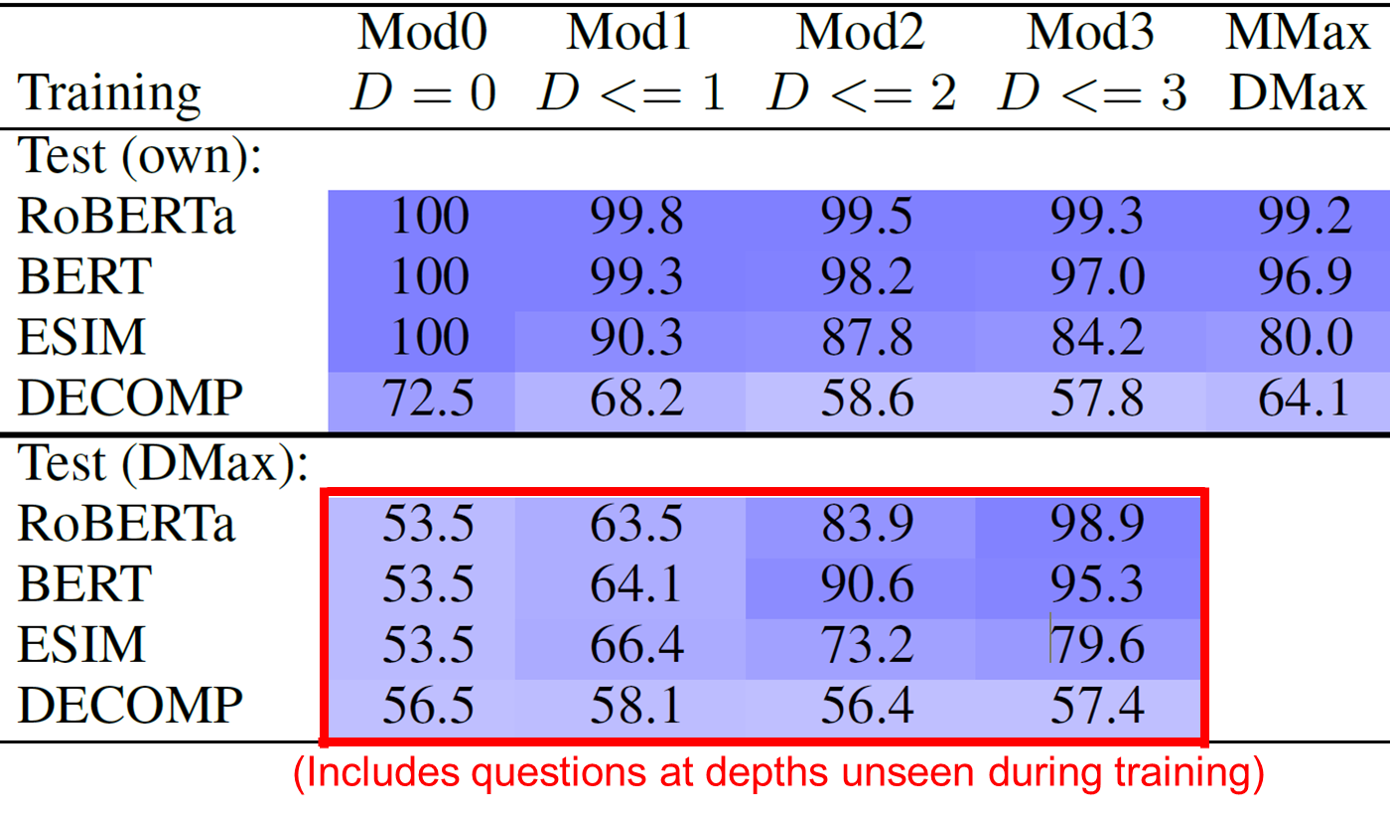}
\eat{
{\small
\setlength{\tabcolsep}{3pt}	% narrower columns
\begin{tabular}{lccccc}
\hline
 & Mod0 & Mod1 & Mod2 & Mod3 & \MMax \\
Training & $D=0$ & $D<=1$ & $D<=2$ & $D<=3$ & \DMax\\\hline
Test (own): &  &  &  &  & \\
RoBERTa & \cellcolor{blue!50} 100 & \cellcolor{blue!50} 99.8 & \cellcolor{blue!50} 99.5 & \cellcolor{blue!50} 99.3 & \cellcolor{blue!50} 99.2\\
BERT & \cellcolor{blue!50} 100 & \cellcolor{blue!50} 99.3 & \cellcolor{blue!49} 98.2 & \cellcolor{blue!48} 97.0 & \cellcolor{blue!48} 96.9\\
ESIM & \cellcolor{blue!50} 100 & \cellcolor{blue!45} 90.3 & \cellcolor{blue!44} 87.8 & \cellcolor{blue!42} 84.2 & \cellcolor{blue!40} 80.0\\
DECOMP & \cellcolor{blue!38} 72.5 & \cellcolor{blue!30} 68.2 & \cellcolor{blue!25} 58.6 & \cellcolor{blue!25} 57.8 & \cellcolor{blue!28} 64.1 \\\hline

\hline
Test (DMax): &  &  &  &  & \\
RoBERTa & \cellcolor{blue!27} 53.5 & \cellcolor{blue!32} 63.5 & \cellcolor{blue!42} 83.9 & \cellcolor{blue!49} 98.9 & \\
BERT & \cellcolor{blue!27} 53.5 & \cellcolor{blue!32} 64.1 & \cellcolor{blue!45} 90.6 & \cellcolor{blue!48} 95.3 & \\
ESIM & \cellcolor{blue!27} 53.5 & \cellcolor{blue!33} 66.4 & \cellcolor{blue!37} 73.2 & \cellcolor{blue!40} 79.6 & \\
DECOMP & \cellcolor{blue!25} 56.5 & \cellcolor{blue!26} 58.1 & \cellcolor{blue!25} 56.4 & \cellcolor{blue!25} 57.4 & \\
\hline
\end{tabular}
}
} % end eat
\vspace{-7mm}
\caption{Transformers (RoBERTa,BERT) are sufficient but not strictly necessary for this task, although
other architectures (ESIM) do not score as well. DECOMP was run as a sanity check that the
datasets are not trivially solvable - its low score (random baseline is 50\%) suggests they are not.}
\label{other-architectures}
\end{table}

We observe that the strongest BERT model trained up to depth 3 (Mod3) masters
the dataset that includes higher inference depths (\DMax)
with 95\%+ accuracy, while ESIM's scores are lower ($\approx$80\%). Note that
unlike RoBERTa and BERT, ESIM was not pre-trained on large amounts of text,
perhaps contributing to its lower scores. This suggests that our results are
not specific to RoBERTa or transformers, although transformers seem to learn the
tasks more easily. As expected, DECOMP does not do well (a random baseline score is 50\%),
suggesting the datasets are not trivially solvable.

A related question is: how important is pretraining? To test this, we
generated a version of the D$\leq$3 dataset in which every word was (systematically)
replaced by a random word, so that there was no grammaticality in the theories.
After training (using the best hyperparameter settings we could find, using the
dev partition), RoBERTa scores 83.3\% on the test partition, substantially below
the 99.3\% on the original dataset in restricted English (Table~\ref{main-results}).
This suggests that the pretrained knowledge in RoBERTa is playing a role in
its performance and making learning easier. Similarly the zero-shot transfer to hand-authored language,
Figure~\ref{nlp-example} and Table~\ref{natlang}, suggests pretrained
knowledge of language may be playing a role.

\section{Discussion and Future Work \label{discussion}}

Although our demonstrations have been in a limited setting,
the implications of being able to predictably reason with
language are significant. With further advances, we may potentially be able to:
\begin{ite}
\item author theories in English (e.g., Figure~\ref{electricity}), thus sidestepping the intricacies of
formal languages and offering new opportunities for easy creation and maintenance of knowledge.
\item have the machine apply {\it general} knowledge, e.g., from Wikipedia, to explainably solve novel problems
\item teach our AI when it makes a mistake, by providing the missing facts and/or correcting the erroneous ones it
used (``instructable systems'').
\item reason about counterfactual situations. For example, we might describe a world in which
plastic is a type of metal, and see how the conductivity of objects change. This useful capability has
previously been out of scope for transformers. % , as they have been tethered to their pretrained knowledge.
\end{ite}
Our RuleTaker models demonstrate
% We have demonstrated
these capabilities in a narrow setting. We now discuss additional
steps needed to achieve these goals more broadly.

\subsection{Extending The Theory Language \label{extending-the-language}}

While we have shown that transformers can emulate a form of deductive reasoning,
our demonstrations have been with small theory sizes ($<$ 20 facts, $<$ 10 rules),
small domains ($<$ 100 possible ground facts), and with a limited rule language
(at most one variable that is universally quantified over). Expanding the
expressiveness of the rule language would enhance the model's 
utility. For example, we have not yet explored using multi-variable rules
such as ``If a person's father is a second person, and the second person's father
is a third person, then the first person's grandfather is the third person,''
limiting what can be stated (e.g., rules of transitivity). Similarly there are
other forms of reasoning we would like to train the model to handle, e.g.,
taxonomic inheritance, reasoning with disjunctive conclusions,
and handling functional relations (``A country has exactly one capital'').
\eat{
other common forms of knowledge we would like to train the model to
reason with, e.g.,
\begin{ite}
\item Taxonomic knowledge (``Dogs are mammals.'')
\item Functional relations (``A country has exactly one capital'')
\item Cardinality (``Mars has two moons'')
\end{ite}
}
This again requires characterizing the semantics of such statements, and
generating training data showing the valid conclusions.

More generally, there are many natural language statements whose formal
meaning is less clear (e.g., ``Most birds fly'', ``It often rains in Seattle in winter.'').
To apply our methodology to statements with more complex semantics would require new training data, either
synthesized from a richer formal representation and model of inference,\footnote{
If one even exists - formal reasoning is still far from modeling all
of natural language inference.} or collected from people.
% In addition, we do not yet know if more complex semantics is learnable by transformers.
% Note that 
% it is unclear if transformers can learn more complex semantics
% despite the results presented here.

\subsection{Generating Training Data \label{generating-training-data}}

We assume that our synthetic training data is sufficiently representative
of the real problems that the model will eventually be used for. However,
it is possible that the generation procedure under-represents or misses
some important types of theory, potentially giving the model a ``blind spot''
on novel problems if it is unable to fully generalize. (A minor example of this was the
MMax results on Electricity4, last paragraph of Section~\ref{hand-authored}).
It would be valuable to find ways to characterize the different {\it types} of inference
problems in the space, and design training curricula to ensure they are systematically
covered and/or the model is able to generalize to them.
Adversarial approaches to generation, where the generator learns to create
theories that are hard for a partially trained model, may be useful in this
context, e.g., \cite{programming-puzzles,Goodfellow2016NIPS2T}.

% Consider the space of possible theories, where each point in the space denotes
% a different theory. The theory generation procedure, used for training,
% samples points in this space, and can be viewed as defining a probability
% distribution over that space. 

\subsection{Formal Theorem Proving \label{theorem-proving}}

If a transformer can reliably emulate the results of correct reasoning, it may have
utility for the formal reasoning community. In particular, if its output
generalizes to more complex problems than seen during training (as our
experiments demonstrate for one particular setting, Table~\ref{main-results}),
one might be able to train on problems that are solvable with reasonable computing
resources, but then apply it to more complex problems that are computationally
infeasible, and still obtain high accuracy. Even if accuracy is not perfect, the
results might help inform more rigorous theorem-proving. For example, one could
imagine a transformer as a ``proof assistant'' that finds likely true conclusions
that could then be verified by a formal reasoner (where verification is easier
than proof from scratch). Our results in Section~\ref{explanation}
suggest first steps to reconstructing proofs using the facts and
rules in the theory. Similarly, a system that can identify
facts that are likely true may help guide model generators,
e.g., \cite{Niemel1997SmodelsA}.

Although we have presented theories in English, another open question
is whether transformers could reason with theories expressed in their
native (formal) form, rather than translated. Similarly, we
have not investigated the transformer's behavior with inconsistent
theories, again a potential avenue for further work.

% Even though our model is not a theorem prover, in the sense of
% ``producing new inference steps using rules of inference.''\footnote{
%  https://en.wikipedia.org/wiki/Automated\_theorem\_proving},
%  it emulates the truth assignments of the theorem prover used to generate
% the training data
% suggest first steps to reconstructing proofs based on rules of inference.
%  Strictly speaking our model is not a theorem prover, in the sense of
% ``producing new inference steps using rules of inference.''\footnote{
% https://en.wikipedia.org/wiki/Automated\_theorem\_proving}. 
% Rather, it emulates the truth assignments of the theorem prover used to generate
% the training data, although the results in Section~\ref{explanation}
% suggest first steps to reconstructing proofs based on rules of inference.
\eat{
The theorem prover that generated our training data follows
the semantics of logic programs, in particular using
negation as failure and making the CWA, and is complete
with respect to consistent, stratified, propositional programs.
Thus our transformer is emulating complete reasoning with
respect to these semantics. If we interpret our theories
with classical logic semantics, though, our theorem prover
(hence the transformer) is incomplete,\footnote{
For example, the theory \{$a \rightarrow b$\} has three (formal) models,
\{$\neg a,\neg b$\}, \{$a,b$\}, and \{$\neg a,b$\} but our theorem
prover (deliberately) only finds the first, as $a$ is not supported
(disallowed by the the semantics of logic programs \cite{Apt1988TowardsAT}.}.
}

Note that as we increase the complexity/size of the theory,
we reach the limit on the number of tokens a transformer can handle ($\leq$512).
Thus handling larger theories might also require new transformer architectures,
or new methods for fetching relevant rules to reason with from a larger knowledge base.

\eat{
\subsection{Models of Inference \label{models-of-inference}}

% We have shown that our trained model emulates reasoning, but 
% which particular model of reasoning is it emulating?
Our training data is generated using a particular semantic theory
of inference, but alternative semantics are possible. For example,
we make a closed-world assumption (CWA) that anything unproven is false, and
(following the semantics of logic programs \cite{Apt1988TowardsAT}) we require true
facts to be {\it supported}.
%\footnote{
%A fact is supported if it is either known, or is the conclusion of a rule
%whose body is true. Requiring support provides a mechanism for resolving
%ambiguities that would otherwise exist. For example, the theory
%\{$\neg a \rightarrow b$\} has three possible models \{$\neg a,b$\},
% \{$a,\neg b$\}, and \{$a,b$\}, but the latter two are disallowed by
%the semantics of logic programs as $a$ is unsupported by the theory.}
However, these are just design decisions, and other alternatives
are possible. For example, we could have made an
open world assumption that unproven statements have an unknown truth
value, and trained the system using examples generated with a three-valued logic.
Or, we could use an inference model that handles uncertainty, e.g., admitting linguistic
modifiers such as ``often'', ``sometimes'', ``a few'', etc. and train a transformer
with examples from that.
}

\subsection{Natural Language Inference (NLI) \label{nli}}

We have shown that transformers can perform deductive inference over English
statements. % , using a particular semantic theory of deduction (namely, logic programs).
However, human reasoning over language - natural language
inference (NLI) - is not always deductive. In particular, NLI allows for
{\it unsupported} inferences that ``a person would typically infer'' \cite{Dagan2013RecognizingTE},
while we have used a precise model of inference in which {\it all} of a rule's conditions
need to be proven true in order for the conclusion to follow. Our model
may still be quite far from that required for fully natural reasoning over language.
For example, we would like our model to still proceed if there are gaps in
the explicitly provided knowledge, providing the missing knowledge is ``obvious'' (and not contradicted
by the explicitly provided facts), perhaps by leveraging its pretrained knowledge.
Similarly, our model's treatment of negation as failure (NAF) sometimes clashes with
intuitions about NLI, for example given (just) ``If my car does not have gas then it is not working.''
our model will conclude (given nothing else) that ``My car is not working.'' as it
cannot {\it prove} that ``My car has gas.''.

This raises a fundamental tension about the nature of the reasoning we ultimately desire:
We want reasoning to be rigorous (conclusions justified by the information provided),
but also ``soft'' (tolerant of phrasing differences and commonsense knowledge gaps), and
strictly speaking these two goals are in conflict. Our experiments with Turk-authored
language illustrates tolerance of phrasing differences, which we view as desirable,
although in a strict deductive sense it is unjustified to conclude
(say) ``A person is green'' from ``Charlie has green teeth''
(Figure~\ref{nlp-example}). Similarly we would like the model
to tolerate minor, unstated taxonomic gaps, for example given
``Buildings have roofs'' conclude ``My house has a roof'', 
even if ``Houses are buildings'' is not explicitly stated
(but {\it not} conclude that result if it is explicitly stated
that ``Houses are {\it not} buildings'').
% \footnote{
% Similarly, the hand-authored explanations for 4th Grade Science questions in \cite{Jansen2018WorldTreeAC}
% include many subtle gaps from a logical point of view, yet still seem reasonable to a human reader.}
Characterizing which inferences should
be deductive vs. which can be assumed in NLI, and training a model
to combine explicitly stated knowledge with implicit (pretrained) knowledge,
remain significant open challenges.
% Regardless of how these
% issues are resolved, though, our experiments suggest that transformers may 
% may be able to learn the target reasoning given appropriate training.\footnote{
% In principle, existing datasets for NLI already contain such examples.
% In practice, though, most NLI and entailment datasets have largely focused on
% linguistic equivalence/subsumption between sentences, rather than
% the forming chains of reasoning.}

% \subsection{Formal Theorem Proving}

% \vspace{-1mm}
\section{Conclusion \label{conclusion}}

% Can transformers emulate reasoning with rules (``advice'') in English? We
% have presented substantial evidence that they can, at least within the constrained
% environments explored in this paper, including reasoning at greater inference
% depths than seen in training, on independently hand-crafted problems, and
% problems expressed in more natural language.

Just as McCarthy advocated 60 years ago for machines reasoning (``taking advice'')
in logic, we have shown (in a restricted setting) that machines can by trained
to reason over language. While we have assumed a particular semantics of
inference, the methodology we have used is general: Characterize the
desired behavior in a formal way, synthesize examples, generate linguistic
equivalents, and train a model. The result, at least within our experiments,
appears to be both natural and robust, in a way distinct from working
with the original formalization.

The ability to reason (or emulate reasoning) over rules expressed in language
has potentially far-reaching implications. For example, rules might be
easily authored by a person, sidestepping some of the intricacies of
a formal language (a simple kind of ``programming in English''); or
they could be retrieved from natural sources (e.g., science texts, Wikipedia).
% For example, rules might be retrieved
% from natural sources (e.g., science texts, Wikipedia), and if a system's answer is derivable from a subset
% of retrieved rules, then those rules form a valid explanation for
% the answer.
Similarly, if the answer is wrong, the user may be able to
directly teach the system by providing general missing
knowledge (or correcting erroneous knowledge) that
can then also be used for new problems - a step
towards instructable algorithms.
% This is made possible
% because the model is behaving in a predictable, rational,
% and hence controllable way with what it knows.
Finally, the mechanism opens the door to
% Similarly, if the answer is wrong, the user may be able to
% directly teach the system by identifying the erroneous/missing
% knowledge and correcting it, again made possible because
neural counterfactual reasoning. For example, we can modify the
earlier ``birds'' rulebase to describe a world in which birds typically
don't fly, but where ostriches can fly, and see the consequences;
or imagine a world in which plastic is a type of metal, and see
how the conductivity of objects change (Appendix~D). 
To encourage further progress, an interactive demo and
all our datasets are available at \demourl~and \dataurl.

\eat{
to reason with. For example, we can modify the
earlier ``birds'' rulebase to describe a world in which birds typically
don't fly, but where ostriches can fly, and see the consequences.
% or imagine a world in which plastic is a type of metal, and see
% how the conductivity of objects change. %  (Appendix~C). 
To encourage further progress, all our datasets and an
interactive demo are available at \demourl.
% We hope that such exciting possibilities will be explored in the near future.
}

% \vspace{2mm}
% \noindent
% {\bf Acknowledgements:}
\section*{Acknowledgements}
Thanks to Chitta Baral, Jonathan Berant, Oren Etzioni, Matt Gardner, Ashish Sabharwal, and Alon Talmor
for comments on earlier drafts.

\bibliographystyle{named}
\bibliography{references}

\newpage

{\centering
{\LARGE \bf Appendix}
}

\section*{A. Generating Theories}

Theories are built with a simple grammar, as follows. Predicates $r(x,y)$ are expressed in infix notation $(x~r~y)$ plus an extra polarity argument denoting if they are negated or not.
\vspace{1mm}
{\small
% \fbox{%
%   \parbox{0.85\columnwidth}
\begin{Verbatim}[frame=single]
   theory = statement+
   statement = fact | rule

   fact =   ( entity "is" attribute polarity )
          | ( entity relation entity polarity )
   rule = ( (fact+) -> fact )
   
   entity = name | variable
   
   name = "Anne" | "Bob" | "Charlie" ...
   variable = "something" | "someone" | "thing"
   
   attribute = "red" | "kind" | "nice" ... 
   relation = "likes" | "chases" | "eats" ...
   polarity = "+" | "-"
\end{Verbatim}
}

\noindent
(, ), and -$>$ denote symbols that appear in the final facts and rules.
``polarity'' denotes whether the fact is negated (``-'') or not (``+'').
The three alternative variable names are simple synonyms, for
variability in language generation.
An example theory (Type1, attributes with negation) following this grammar is:

\vspace{1mm}
{\small
\begin{Verbatim}[frame=single]
("Bob" "is" "big" "+")       // Bob is big.
("Bob" "is" "green" "-")     // Bob is not green.
("Bob" "is" "quiet" "-")     // Bob is not quiet.

// Nice, smart people are rough.
((("someone" "is" "nice" "+")
  ("someone" "is" "smart" "+"))
        -> ("someone" "is" "rough" "+"))

// If someone is quiet and round then they are
((("someone" "is" "quiet" "+")        //  not big.
  ("someone" "is" "round" "+"))
        -> ("someone" "is" "big" "-"))

// If Bob is not green then Bob is nice.
((("Bob" "is" "green" "-"))
        -> ("Bob" "is" "nice" "+")))
\end{Verbatim}
}

\noindent
We now provide details of how the generation is controlled.

\begin{figure*}
{\small \begin{verbatim}
((("something"
((("something" "loves"                                                                    // H1
((("something" "loves" "cat"                                                              // H2
((("something" "loves" "cat" "+") ("cat"
((("something" "loves" "cat" "+") ("cat" "is"                                             // H3
((("something" "loves" "cat" "+") ("cat" "is" "happy"                                     // H2
((("something" "loves" "cat" "+") ("cat" "is" "happy" "-"))                               // H2
((("something" "loves" "cat" "+") ("cat" "is" "happy" "-")) ->
((("something" "loves" "cat" "+") ("cat" "is" "happy" "-")) -> ("something"
((("something" "loves" "cat" "+") ("cat" "is" "happy" "-")) -> ("something" "loves"       // H3
((("something" "loves" "cat" "+") ("cat" "is" "happy" "-")) -> ("something" "loves" "dog" // H2
((("something" "loves" "cat" "+") ("cat" "is" "happy" "-")) -> ("something" "loves" "dog" "+"))
\end{verbatim}}
  \caption*{Figure A1: The incremental generation of a rule, illustrating where the heuristics H1-H3 (Appendix~A.4) are applied during generation.
    The final rule would be rendered in English as (for example):
    {\it If something loves the cat and the cat is not happy then it loves the dog.}}
  \end{figure*}

  \subsection*{A.1 Names, Attributes, and Relations}

  As described in Section~\ref{theory-generation}, we generate two types of
  theories, Type1 (denoted ``Att'') only contains attributes, and uses people names as entity names.
  Type2 (denoted ``Rel'') contains both attributes and relations, and uses animal names as entity names.

  Specifically, we use the following {\it pools} of names, attributes, and relations.
  \vspace{1mm}
  \noindent
For the Type1 {\it attribute} ("Att") theories:
\begin{des}
\item[{\bf name-pool:}] "Anne" "Bob" "Charlie" "Dave" "Erin" "Fiona" "Gary" "Harry" 
\item[{\bf attribute-pool:}] "red" "blue" "green" "kind" "nice" "big" "cold" "young" "round" "rough" "white" "smart" "quiet" "furry" 
\item[{\bf relation-pool:}] (none)
\end{des}
\vspace{1mm}
For the Type2 {\it relational} ("Rel") theories:
\begin{des}
\item[{\bf name-pool:}] "cat" "dog" "bald eagle" "rabbit" "mouse" "tiger" "lion" "bear" "squirrel" "cow"  
\item[{\bf attribute-pool:}] "red" "blue" "green" "kind" "nice" "big" "cold" "young" "round" "rough" 
\item[{\bf relation-pool:}] "likes" "chases" "eats" "sees" "visits" "needs"
  \end{des}

For each new theory, we use just a {\it random subset} of names, attributes, and relations from these pools
to generate the theory. We do this to ensure that there's a reasonable chance of repeated
names/attributes/relations in the theories, to better mimic real theories. The following
parameters control the size of these subsets:
\begin{ite}
\item \#names: randomly select a number from 2-4  
\item \#attributes: randomly select a number from 1-5 
\item \#relations:
  \begin{ite}
  \item 0 (for Type1 Att theories)
  \item randomly select from 1-4 (Type2 Rel theories)
  \end{ite}
\end{ite}

  \subsection*{A.2 Theory Size during Generation}

The theory size is similarly controlled by parameters randomized for each theory:
\begin{ite}
\item \#rules: select a random number from 1-8 
\item \#facts: select a random number from 1-16
\end{ite}

\subsection*{A.3 Controlling Paths through the Grammar}

Two grammar rules use non-uniform transition probabilities as follows
\begin{ite}
\item {\tt fact} rule: 
 \begin{ite}
 \item Type1 (``Att'') theories: Always select "is"
 \item Type2 (``Rel'') theories: select "is" with p=0.3, relation with probability p=0.7 
 \end{ite}  
\item {\tt polarity} (negation) rule: 
 \begin{ite}
 \item theories without negation ("NoNeg"): always select "+"
 \item theories with negation ("Neg"): select "+" with p=0.8, "-" with p=0.2
 \end{ite}
\end{ite}

\subsection*{A.4 Special Heuristics for Rule Generation \label{heuristics}}

The following heuristics are to encourage the generator to produce rules that are more realistic (with shared arguments) rather than just randomly chosen entities. They encourage shared arguments (heuristic H3), occasionally fully grounded rules (heuristic H1), ensures that there are no free variables in the rule's conclusion (heuristic H3), and avoids multi-variable rules (heuristic H2). Multi-variable rules were avoided due to the challenge of rendering them fluently in English, but would be a valuable extension to explore further (Section~\ref{extending-the-language}). Similarly, these heuristics could likely be improved to better sample the space of realistic theories, or the generator replaced with an adaptive one (Section~\ref{generating-training-data}).

\vspace{1mm}
\noindent
When generating a rule: 
\begin{ite}
\item \#rule-conditions: for each rule, select a random number from 1-2
\end{ite}

\noindent
Special rule for entity selection during rule generation:
\begin{des}
\item[H1.] For the {\it first} entity (arg1) of the {\it first} fact in the rule condition:
\begin{ite}
  \item select a (random) name with p=0.2, a variable with p=0.8
\end{ite}  
\item[H2.] For each {\it second} entity (arg2) of all the rule's conditions and conclusion, i.e., when a relational fact is being built (which takes an entity as arg2):
\begin{ite}
 \item always select a (random) name (never a variable)
 \end{ite}
\item[H3.] For the {\it first} entity (arg1) of {\it all subsequent} entity choices in the rule's condition and conclusion:
\begin{ite}
 \item select an {\it already used} name or variable from the earlier parts of the rule so far
\end{ite}
\end{des}

\noindent
An example of the generation sequence for a rule, showing how the above heuristics are applied, is shown in Figure~A.

\subsection*{A.5 Ensuring Reasoning Depth}

Depth of reasoning is ensured using naive generate-and-test:
Generate a theory, derive all implications using the reasoner (Section~\ref{forward-inference}), and test
that the deepest depth of reasoning used is at least as deep as the target depth $D$.
If it is not, discard the theory and repeat.

\section*{B. Logic for the `Birds'' Rulebase}

\noindent
Sergot's original logic\footnote{https://www.doc.ic.ac.uk/$\sim$mjs/teaching/KnowledgeRep491/ ExtendedLP\_491-2x1.pdf, p5}
for the ``birds'' rulebase is as follows (``not'' denotes negation as failure, ``$\neg$'' denotes hard negation):
\begin{myquote}
{\it  \% Rules:} \\
{\it \% If someone is a bird and not abnormal then they can fly.} \\    
can\_fly(X) $\leftarrow$ bird(X), not abnormal\_bird(X) 
\vspace{1mm} \\
{\it \% If someone is an ostrich then they are a bird.}  \\
bird(X) $\leftarrow$ ostrich(X) 
\vspace{1mm} \\
{\it \% If someone is an ostrich then they are abnormal.} \\
abnormal\_bird(X) $\leftarrow$ ostrich(X) 
\vspace{1mm} \\
{\it \% If someone is an ostrich then they cannot fly.} \\
$\neg$can\_fly(X) $\leftarrow$ ostrich(X) 
\vspace{1mm} \\
{\it \% If someone is a bird and wounded} \\
{\it \% \hspace*{1mm} then they are abnormal.} \\
abnormal\_bird(X) $\leftarrow$ bird(X), wounded(X) 
\vspace{1mm} \\
{\it \% If someone is wounded then they cannot fly.} \\
$\neg$can\_fly(X) $\leftarrow$ wounded(X) \\
\ \\
{\it  \% And the following facts:} \vspace{1mm} \\
{\it \% Arthur is a bird and not wounded.} \\
bird(arthur). $\neg$wounded(arthur). \vspace{1mm} \\
{\it \% Bill is an ostrich.} \\
ostrich(bill). \vspace{1mm} \\
{\it \% Colin is a bird and wounded.} \\
bird(colin). wounded(colin). \vspace{1mm} \\
{\it \% Dave is not an ostrich and wounded.} \\
$\neg$ostrich(dave). wounded(dave). 
\end{myquote}

\section*{C. The ``Electricity'' Rulebases}

The four electricity rulebases, and the scenario vocabularies, exactly as
provided to the model, are shown below: 

\subsection*{C.1 Electricity1}

\subsubsection*{Rulebase}

\noindent % \begin{tt}
If a circuit has a switch and the switch is on then the circuit is complete. \\
If a circuit does not have a switch then the circuit is complete. \\
If a circuit is complete and the circuit has a light bulb then the light bulb is glowing. \\
If a circuit is complete and the circuit has a bell then the bell is ringing. \\
If a circuit is complete and the circuit has a radio then the radio is playing. 
% \end{tt}

\subsubsection*{Scenario Generation}

\noindent % \begin{tt}
A circuit has a switch. {\it (included as a fact with p=0.5)} \\
A switch is on. {\it (p=0.5)} \\
A circuit has a light bulb. \vt~ A circuit has a bell \vt~ A circuit has a radio. {\it (select 1)} 
% \end{tt}

\subsection*{C.2 Electricity2}

\subsubsection*{Rulebase}

\noindent % \begin{tt}
If a circuit has a switch and the switch is on then the circuit is complete. \\
If a circuit does not have a switch then the circuit is complete. \\
If a circuit is complete then a current runs through the circuit. \\
If a current runs through a circuit and the circuit has a light bulb then the light bulb is glowing. \\
If a current runs through a circuit and the circuit has a bell then the bell is ringing. \\
If a current runs through a circuit and the circuit has a radio then the radio is playing. 
% \end{tt}

\subsubsection*{Scenario Generation}

\noindent % \begin{tt}
A circuit has a switch. {\it (p=0.5)} \\
A switch is on. {\it (p=0.5)} \\
A circuit has a light bulb. \vt~ A circuit has a bell \vt~ A circuit has a radio. {\it (select 1 fact)} 
% \end{tt}

\subsection*{C.3 Electricity3}

\subsubsection*{Rulebase}

\noindent % \begin{tt}
If a circuit has a battery then the circuit is powered. \\
If a circuit does not have a battery then the circuit is dead. \\
If a circuit is dead then a bell is not ringing. \\
If a circuit is dead then a radio is not playing. \\
If a circuit is dead then a light bulb is not glowing. \\
If a circuit has a switch and the switch is on then the circuit is complete. \\
If a circuit does not have a switch then the circuit is complete. \\
If a circuit is powered and the circuit is complete then a current runs through the circuit. \\
If a current runs through a circuit and the circuit has a light bulb then the light bulb is glowing. \\
If a current runs through a circuit and the circuit has a bell then the bell is ringing. \\
If a current runs through a circuit and the circuit has a radio then the radio is playing. 
% \end{tt}

\subsubsection*{Scenario Generation}

\noindent % \begin{tt}
A circuit has a battery. {\it (p=0.9)} \\
A circuit has a switch. {\it (p=0.5)} \\
A switch is on. {\it (p=0.5)} \\
A circuit has a light bulb. \vt~ A circuit has a bell \vt~ A circuit has a radio. {\it (select 1)}
% \end{tt}

\subsection*{C.4 Electricity4}

\subsubsection*{Rulebase}

\noindent % \begin{tt}
If a circuit includes a battery and the battery is not flat then the circuit is powered. \\
If a circuit includes a switch and the switch is on then the circuit is complete. \\
If a circuit does not include a switch then the circuit is complete. \\
If a wire is metal then the wire is conducting. \\
If a wire is plastic then the wire is not conducting. \\
If a circuit is powered and the circuit is complete and a wire is conducting then a current runs through the circuit. \\
If a current runs through a circuit and the circuit includes a light bulb then the current runs through the light bulb. \\
If a current runs through a circuit and the circuit includes a bell then the current runs through the bell. \\
If a current runs through a circuit and the circuit includes a radio then the current runs through the radio. \\
If a current runs through a light bulb then the light bulb is glowing. \\
If a current runs through a bell then the bell is ringing. \\
If a current runs through a radio then the radio is playing. 
% \end{tt}

\subsubsection*{Scenario Generation}

\noindent % \begin{tt}
A circuit includes a battery. {\it (p=0.9)} \\
A battery is flat. {\it (p=0.2)} \\
A circuit includes a switch. {\it (p=0.5)} \\
A switch is on. {\it (p=0.7)} \\
A wire is metal. \vt~ A wire is plastic. {\it (select 1, p(metal)=0.9)} \\
A circuit includes a light bulb. \vt~ A circuit includes a bell \vt~ A circuit includes a radio. {\it (select 1)} 
% \end{tt}

\section*{D. Counterfactuals}

Because our models reason with explicitly provided knowledge
(rather than latent, fixed, pretrained knowledge), they can
support counterfactual reasoning by providing counterfactual
statements in the context. This is something that transformers have
not previously been used for. As an anecdotal example, we take
a context that implies that metal nails are conductors (where C=context, Q=question, A=answer),
and illustrate with our model's predictions\footnote{Live demo available at \demourl}:
\begin{description}
\item[{\bf C:}] \it{Metal things conduct electricity. Insulated things do not conduct electricity. Nails are metal.}
\item[{\bf Q:}] \it{A nail conducts electricity?}
\item[{\bf A:}] {\tt TRUE}
\end{description}
We can now make this counterfactual by declaring metal as an insulator. The model
then correctly changes its answer to conclude metal nails would then not conduct
electricity:
\begin{description}
\item[{\bf C:}] \it{Metals conduct electricity. Insulators do not conduct electricity. \underline{Nails are insulators.}}
\item[{\bf Q:}] \it{A nail conducts electricity?}
\item[{\bf A:}] {\tt FALSE}
\end{description}
Similarly given a context implying plastic nails do not conduct:
\begin{description}
\item[{\bf C:}] \it{Metals conduct electricity. Insulators do not conduct electricity. \underline{Plastic is an insulator. Nails are plastic.}}
\item[{\bf Q:}] \it{A nail conducts electricity?}
\item[{\bf A:}] {\tt FALSE}
\end{description}
we can modify it by adding the counterfactual that plastics are metals. Again the model
correctly changes its answer to infer that therefore plastic nails will conduct electricity:
\begin{description}
\item[{\bf C:}] \it{Metals conduct electricity. Insulators do not conduct electricity. \underline{Plastic is a metal.} Nails are plastic.}
\item[{\bf Q:}] \it{A nail conducts electricity?}
\item[{\bf A:}] {\tt TRUE}
\end{description}
% Similarly for the ``Birds'' rulebases (Figure~\ref{birds}), we might consider
% a counterfactual world in which birds are typically flightless, but
% ostriches can fly, and see the consequences.
This capability is a potentially powerful new avenue for using transformers.

\end{document}